\documentclass[conference]{IEEEtran}
\usepackage{hyperref}
\usepackage[dvipsnames]{xcolor}

\hypersetup{
	colorlinks   = true,
	citecolor    = blue,
	linkcolor    = red
}
\usepackage{subfigure}
\usepackage{amsmath}
\usepackage{graphicx}
\usepackage{tabularx}
\usepackage{xcolor}
\usepackage{multirow}
\usepackage[numbers]{natbib}

\ifCLASSINFOpdf
\else
\fi
\usepackage{algorithmic}
\usepackage{algorithm}

\begin{document}
%
\title{Eavesdrop the Composition Proportion of Training Labels in Federated Learning}

\author{\IEEEauthorblockN{Lixu Wang}
\IEEEauthorblockA{Zhejiang University\\
wanglixu\_eecs@zju.edu.cn}
\and
\IEEEauthorblockN{Shichao Xu}
\IEEEauthorblockA{Northwestern University\\
shichaoxu2023@u.northwestern.edu}
\and
\IEEEauthorblockN{Xiao Wang}
\IEEEauthorblockA{Northwestern University\\
wangxiao@cs.northwestern.edu}
\and
\IEEEauthorblockN{Qi Zhu}
\IEEEauthorblockA{Northwestern University\\
qzhu@northwestern.edu}}

\maketitle

\begin{abstract}
Federated learning (FL) has recently emerged as a new form of collaborative machine learning, where a common model can be learned while keeping all the training data on local devices. Although it is designed for enhancing the data privacy, we demonstrated in this paper a new direction in inference attacks in the context of FL, where valuable information about training data can be obtained by adversaries with very limited power. In particular, we proposed three new types of attacks to exploit this vulnerability. The first type of attack, \emph{Class Sniffing}, can detect whether a certain label appears in training. The other two types of attacks can determine the quantity of each label, i.e., \emph{Quantity Inference} attack determines the composition proportion of the training label owned by the selected clients in a single round, while \emph{Whole Determination} attack determines that of the whole training process. We evaluated our attacks on a variety of tasks and datasets with different settings, and the corresponding results showed that our attacks work well generally. Finally, we analyzed the impact of major hyper-parameters to our attacks and discussed possible defenses.
\end{abstract}


%

\newcommand{\shichao}[1]{{\color{purple}(shichao: #1)}}
\newcommand{\note}[1]{{\color{green}(note: #1)}}
\newcommand{\qi}[1]{{\color{red}(qi: #1)}}
\newcommand{\chao}[1]{{\color{blue}(chao: #1)}}
\newcommand{\xiao}[1]{{\color{YellowOrange}(xiao: #1)}}


\section{Introduction}
The emergence of federated learning (FL) enables multiple devices to learn a common model while keeping all the training data on their own devices. It allows for less resource consumption on the cloud and ensures the privacy at the same time.  Multiple applications have benefited from FL, including mobile phones~\cite{anguita2013public,hard2018federated,ramaswamy2019federated}, wearable devices~\cite{pantelopoulos2009survey,nguyen2018d}, autonomous vehicles~\cite{samarakoon2018distributed,samarakoon2018federated}, etc.
In standard federated learning, all participants are required to train their local models. A random subset of clients will be selected each round, who will upload their gradient updates to the central server. Similar FL architectures can be found in~\cite{chilimbi2014project,dean2012large,lin2017deep,moritz2015sparknet,xing2015petuum,zinkevich2010parallelized}.

One interesting question here is about the security and privacy implication in the FL training process. Any characteristic of clients' private data needs to be protected carefully since it may reveal some important private information about the training data -- e.g., the distribution of labels might show the diversity of participants.
Similarly, what the training data consists of is also what attackers want to explore, i.e., can they determine the quantity proportion of different labels in the whole training dataset during the training process? 
This problem may pose serious threats to FL security. For instance, an attacker can acquire information about the morbidity of a particular disease if the government is training an online disease diagnosis system. 
A malicious store can figure out the relation between the supply and demand of a certain product when there is a new commodity registration system trained with FL approach, and it can then adjust its price accordingly to gain unfair advantage. 

In the literature, there are mainly two areas of research on attacking FL models: 
division and aggregation, which correspond to the two main roles in FL (distributed devices and central server). The former one assumes that the attacker compromises some participated devices and uses them to achieve malicious intentions, e.g. importing backdoor to FL~\cite{bagdasaryan2018backdoor,baruch2019little}, adversarial poisoning~\cite{fung2018mitigating,bhagoji2018analyzing,mahloujifar2018multi}, membership or property inference~\cite{melis2018exploiting,truex2019demystifying,pyrgelis2017knock,nasr2018comprehensive}, and reconstruction attack~\cite{hitaj2017deep,salem2019updates,hayes2019logan}. The attacks in the latter area are relatively less-studied. In~\cite{wang2019beyond}, the authors assume that the central server is malicious and train a GAN to reproduce data samples similar to the privacy of clients. Similar idea can also be seen in~\cite{aono2017privacy}. Please see Section~\ref{sec:related} for a more comprehensive discussion on related works.

These previously-studied attacks to FL, e.g., membership inference or reconstruction attack, did not lay much emphasis on the quantity information in training, as they usually focus on existential information, i.e., whether a certain sample exists in training data. 
Another drawback of these approaches is that they all need individual updates that clients sent to the server. However, under the secure aggregation protocol~\cite{bonawitz2017practical} or differential privacy techniques~\cite{geyer2017differentially,mcmahan2017learning}, both the participants and the server cannot acquire the individual updates in the plain form, which make most of these attacks difficult. Therefore, more practical and applicable attacks should be based on the assumption that the observation of individual updates is not available.

The aforementioned issues motivated us to consider attacks without asking individual updates. In this paper, we propose three new inference attacks with high success rate and without the need of any gradient updates from individual clients. In addition, our attacks concentrate on the quantity information of training data in FL, which could lead to serious consequences but has never been studied in prior works to the best of our knowledge. We conducted extensive experiments to evaluate the effectiveness and generality of our approaches, and the results showed the existence of vulnerability from quantity privacy leakage.


\medskip\noindent{\bf Our contributions.} In this paper, we make the first step
towards quantity estimation attacks in federated learning. Specifically:
\begin{itemize}
\item {We propose a new attacking surface in the context of federated learning, i.e., inferring the quantity composition proportion of different labels in the training process. For instance, an attacker may learn how many data samples with each label are used in the training of a certain learning model, which may possibly pose considerable privacy threats to the practical application of FL.}
\item {We design three general attacks towards FL without the need to observe any individual updates. This enables the adversaries to launch our attacks successfully in FL even with secure aggregation protocols or under the protection of differential privacy. Our attacks are passive, which means they will not impose any influence to the training process, and thus they can work covertly without being detected by many intrusion detection techniques.}
\item {Our technique can infer the labels' quantity composition proportion of a single training round, or the whole training process. The former aims at stealing the quantity information of the training data owned by selected clients, while the latter one targets the quantity proportion of all participants at different training stages.}
\end{itemize}

\section{Threat Model}
\label{sec:headings}
\subsection{Problem}
\label{pro}
The advancement of deep learning techniques has received significant interests in recent years. It also has a wide range of applications on different types of devices, which gives a bright stage for FL to show its merits on convenience, privacy protection and resource utilization. In fact, FL has shown great promises not only on 
smartphone-based applications (e.g., human activity recognition~\cite{anguita2013public}, heart rate monitoring~\cite{pantelopoulos2009survey} and keyboard prediction~\cite{hard2018federated,ramaswamy2019federated}), but also in other fields, such as healthcare industry (e.g., disease diagnosis online expert system and medical insurance registration~\cite{huang2018loadaboost}) and transportation systems (vehicular networking technology~\cite{samarakoon2018distributed,samarakoon2018federated}).



FL is designed to preserve the private data of individuals, and any information or characteristic should be protected seriously. 
In FL architecture, training data owned by clients comes from various sources, so the quantity of samples among different labels might be unbalanced, which reflects the clients' overall characteristic. It is a potential source of information leakage if these quantities are illegally obtained by malicious attackers.
For instance, an organization wants to build up an online disease prediction system with FL structure among thousands of hospitals. Each hospital trains their local model with their own data, and the organization will obtain a global model that is able to predict the trend of many diseases, not just a small group of diseases emerged in a single hospital. Then, there is a malicious player who wants to know how many hospitals have treated a particular disease, so that it can raise the corresponding treatment expenses and even estimate the approximate distribution scope of this disease. This is a simple example of attackers may try to learn the composition proportion of training data for their advantage, and many other applications could have similar concern.

Thus, in this work, the goal of attackers is defined as to \textbf{infer the quantity information of particular training labels, especially the composition proportion of training labels in a single training round and the whole training process.}

\subsection{Assumptions}
Unlike prior inference attack, the application setting here is based on more realistic scenarios, i.e., the central aggregation server will choose a set of clients randomly from thousands of participants, which we call the \textbf{selection process}, and collect their gradient updates generated by training local models with each own data in every training epoch. After such collection, we assume a secure aggregation algorithm, which is an important characteristic of FL, is executed so that the server cannot observe individual updates sent by clients in plain text form but can only acquire the aggregated value.

According to the property inference attack \cite{melis2018exploiting,ateniese2013hacking,hitaj2017deep}, we know that particular batches, or particular property of training data, can result in change of gradient on corresponding neurons but have little effect on other neurons. As we know, different training labels are units of different features. Given that we sum up all property inferences towards the feature set for a particular label, is it possible to infer some information about such label rather than just its properties? As we discovered, the answer is yes: an adversary can infer some important information about the training label by analyzing gradient changes in the training process. Here, without loss of generality, we assume that the same labels possessed by different clients result in similar local gradient changes. And if we can determine the global updates consist of how many such local changes, then the number of clients who own the same labels can be obtained.

\begin{figure*}[!t]
\centering
\includegraphics[width=\linewidth]{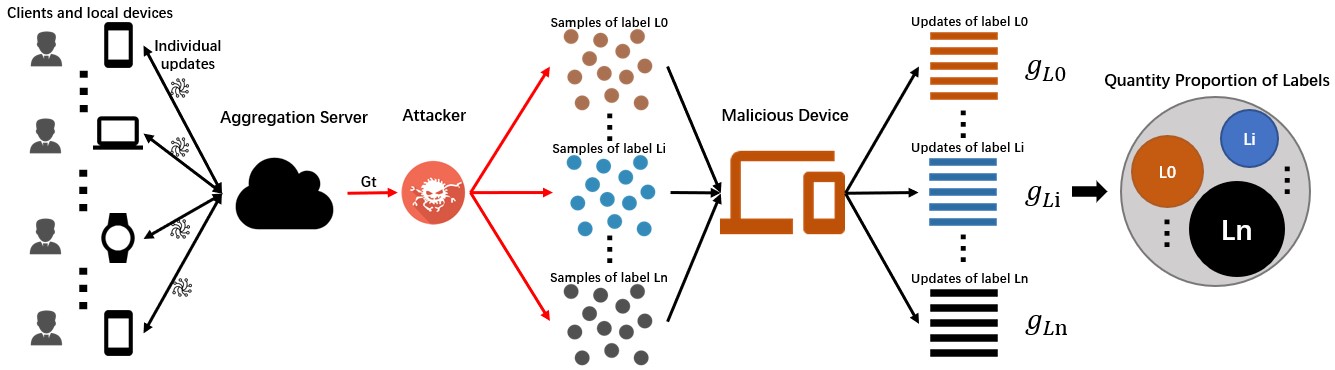}\vspace{-15pt}
\caption{\normalsize {\bf The basic workflow of our inference attacks.} The server collects the gradient updates from selected clients and aggregates them to the current global model $G_t$. The attacker downloads the current global model $G_t$, and trains different labels respectively on the same $G_t$ to obtain corresponding updates $\{g_{L0}, ..., g_{Li}, ..., g_{Ln}\}$. He can then estimate the quantity proportion of labels in the training data by analyzing these updates.
}
\label{fig_workflow}
\end{figure*}

\subsection{Attacker Capacity}
One of the key features of our attacks is that they do not require any observation of individuals' gradient updates, which makes them much easier to be launched than previous attack models. Other basic pre-requirements are similar to other attacks, as discussed below.  

The attacker should obtain some control of a legal participant in FL, specifically, he should be able to acquire complete privileges of reading the content of messages from the aggregation server, comprehending the structure of local model, and modifying or changing the training data with full freedom. He will need some prior knowledge about the training process, i.e., the average number of labels owned by each participant and the probable number of data samples per label. Such information can be estimated by collecting the data of a few participants and performing simple statistical analysis. At last, the attacker should know the approximate number of clients selected by the server in a single training round.

\subsection{Attack Overview}
We propose three original label inference attacks in FL environment:
\begin{enumerate}
    \item \textbf{Class Sniffing.} In a single training round, the adversary is able to infer whether a particular class of training data appears.
    \item \textbf{Quantity Inference.} In a single training round, the attacker can make a judgment that whether a certain training label is owned by a small group of clients or a large group, and predicts how many clients own this label.
    \item \textbf{Whole Determination.} The malicious participant aims to obtain the composition proportion about the dataset labels of current global model.
\end{enumerate}

The inferred information about training labels can be applied to many fields. We list three possible scenarios here:
\begin{enumerate}
\item Use the rare `labels' to identify clients, since these labels are usually owned by extremely few people. Specifically, if such labels are detected in training by our attacks, the attacker can know who participate in the training.
\item Apply this approach to detect the intrusion of malicious participants. The intrusion phenomenon, e.g., backdoor and poisoning attack in FL, was studied in prior works. For instance, \citet{fung2018mitigating} applied cosine similarity to detect Sybil, and \citet{bagdasaryan2018backdoor} proposed a type of powerful backdoor attack under FL scenario. They all mentioned that the updates provided by malicious attackers are different from that of benign clients. Thus, we can regard adversarial data as a type of unique labels that are only owned by malicious clients, and detect them in the training process.
\item Obtain the composition proportion of labels in the training process. We may use some other techniques to train the learning model better (such as data augmentation, focal loss~\cite{lin2017focal}) if we find that the training labels are unbalanced. 
\end{enumerate}

\section{Design}
\subsection{Background}
In supervised learning, we denote the loss function as
\begin{equation}
    Loss_{\theta} = D(F, Y)\label{formula_1}
\end{equation}
where
\begin{equation}
    Y = (f_1(x), f_2(x), \ldots, f_i(x), \ldots, f_k(x))\label{formula_y}\end{equation}
    \begin{equation}
    F = (g_1(x), g_2(x), \ldots, g_i(x), \ldots, g_k(x))\label{formula_F}\end{equation} 

Here $D(\cdot,\cdot)$ could represent distinct formats under different scenarios, e.g., Mean Square Error (MSE) or Mean Absolute Error (MAE). $k$ is the number of label classes. 
$H = (f_1, f_2, \ldots, f_i, \ldots, f_k)$ is the mapping from inputs $x$ to target label $Y=(y_1, y_2, \ldots, y_i, \ldots,  y_k)$; 
$I = (g_1, g_2, \ldots, g_i, \ldots, g_k)$ is the learning model that maps the inputs $x$ to the prediction label $F$. 

The objective of the training process is to minimize the loss of the network, and here we choose the popular stochastic gradient descent (SGD) method to be the network's optimizer. SGD decides how to modify the network parameters $\theta$ in each training iteration. Specifically, it calculates the opposite direction of the gradient of the loss function in terms of every member in $\theta$, combines it with the learning rate $\gamma$, and updates $\theta$ to the next state. When the value of the loss function shrinks to a relative bottom bound, the training process stops. The calculation of gradients is implemented by back-propagation operation from the last to the first layer of the whole network, and the standard updated formula of SGD is $\theta := \theta - \gamma \nabla L(\theta)$.

\subsection{Overview}
The basic process of our attack is presented in Figure~\ref{fig_workflow}. There are one or more observers, in other words, adversarial attackers in the training process. At each training iteration $t$, they download the current global model, which is the detailed parameter information of the network and denoted as $G_t$ (hence FL problems are always in the white-box form). Next, the attackers can train local model with auxiliary dataset to obtain relatively standard gradient changes $g_{t+1} - G_t$, and then conduct analysis between the global updates $G_{t+1} - G_t$ and $g_{t+1} - G_t$ to determine whether a particular label appear in the training round $t$. Furthermore, based on a much deeper comparison between the magnitude of $g_{t+1} - G_t$ and $G_{t+1} - G_t$, the quantity information, i.e., how many clients own a particular training label, can also be acquired. One thing to note is that if these observers/attackers are selected as training clients, their contribution to the global model needs to be removed when comparing the magnitude of $g_{t+1} - G_t$ and $G_{t+1} - G_t$. We name the former type of attack (determining whether a particular label appears) as \textbf{Class Sniffing}, and the latter types (acquiring quantity information) as \textbf{Quantity Inference}. These two types of attack are from the perspective of a single training round. We also propose another label inference attack, \textbf{Whole Determination}, which can determine the composition proportion of training labels in the whole training process. 

\begin{figure}[!t]
\centering
\includegraphics[width=3.2in]{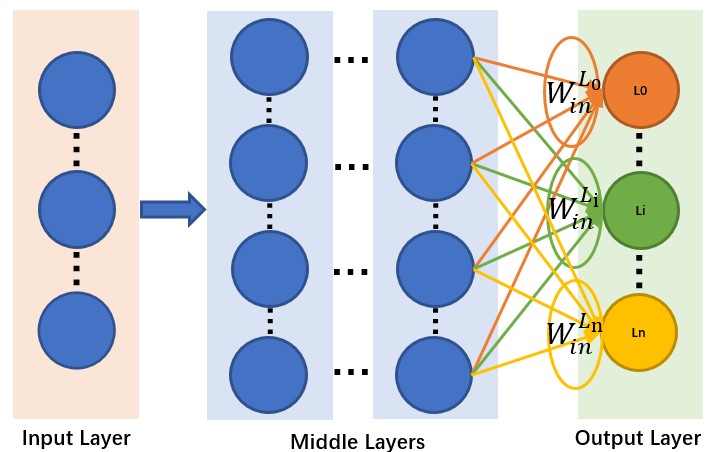}
\caption{\normalsize The positions of the neuron weights we are interested in, denoted as $\{W_{in}^{L0}, \ldots, W_{in}^{Li}, \ldots, W_{in}^{Ln}\}$ of labels $\{L_0, \ldots, L_i, \ldots, L_n\}$, respectively.}
\label{fig_network}
\end{figure}
\subsection{Class Sniffing}
Like most prior work, we build these attack models on the supervised classifying task. We utilize a feed-forward neural network with output size equal to the number of total classes. For each training label, the position of output neurons is shown in Figure~\ref{fig_network}. We discovered a phenomenon that is similar to the basis of property inference attack \cite{ateniese2013hacking}. More specifically, in our experiments, we observed that, using a particular label $L$ in the training will make the inputting weights (the network connection weights denoted as $W_{in}^L$ in Figure~\ref{fig_network}) of corresponding output neuron grow significantly and the weight vectors of other neurons decrease slightly. Such observation motivated our design of the Class Sniffing attack.

We use $N_{int}^L$ to denote the updates of weight set $W_{in}^L$. Both $W_{in}^L$ and $N_{int}^L$ exist in a vector form with the size equal to the number of neurons in the layer before the output layer. For example, when we train a model on the MNIST dataset, the average increase achieves approximately $+3.00$, while the average decrease is $-0.25$. The worst case happens when there is no sample of a particular label in the training data, and then its corresponding inputting weights accept all negative impact without any positive benefit. This case can be simulated with our auxiliary data by restricting this particular label $L$ not to emerge in training, so that the weight updates of its corresponding neurons would be as the worst case. The inputting weight updates vector in such worst case can be regarded as a threshold $Th_{low}$. In a particular round, if the updates of $W_{in}^L$ corresponding to label $L$ are higher than $Th_{low}$, it means that $L$ appears in training; and if the weight changes are approximately close to this threshold, label $L$ can be considered absent in the training round. The detailed acquiring process of such thresholds is shown in Algorithm~\ref{alg:1}.

\begin{algorithm}[!t]
\caption{The threshold acquiring for Class Sniffing}
\label{alg:1}
\begin{algorithmic}
\REQUIRE ~~\\ 
Attacker's auxiliary data samples of different training labels, $D_{L0}, D_{L1},\ldots, D_{Ln}$;\\
Approximate number of labels owned by selected clients in a training round, $N_{L}$;\\
Selection proportion of clients in a training round, $P$;\\
Approximate number of whole participants, $N_{p}$;\\
Inputting weight positions of output layer each label, $W_{in}^{Li}$;
\ENSURE ~~\\ 
The threshold that indicates the existence of a certain label in a training round, $Th_{low}$;\\
\STATE \textbf{Begin:}
\STATE {Receive $\theta_t$ from server} 
\FOR{$i$ = 1 to len($D_{L0}, D_{L1},\ldots, D_{Ln}$)}
\STATE $W$ = []
\FOR{$j$ in $D_{L0}, D_{L1},\ldots, D_{Ln}$}
\STATE $g_j$ = Local\_train($\theta_t$, $j$) - $\theta_t$
\STATE $w$ = acquire($g_j$, $W_{in}^{Li}$) $//$ acquire updates from $g_j$;
\STATE $W$.append($w$)
\ENDFOR
\STATE $W$ = delete($W$, $i$) $//$ delete the updates of $W_{in}^{Li}$ on $g_i$;
\STATE $Th_{low}^i$ = $N_{p}$ * $P$ * $N_{L}$ * mean($W$)
\ENDFOR

Local\_train($\theta_t$, $D_{L}$):
\STATE $\quad$ use $D_L$ to train local model $\theta_t$
\STATE $\quad$ \textbf{return} local model $g_{t+1}$
\end{algorithmic}
\end{algorithm}

\subsection{Quantity Inference}
\label{qi}
Similar to the workflow of Class Sniffing, in Quantity Inference, malicious attacker trains his local model using auxiliary data, especially just using data samples of a single label, and then obtain several local updates $\{g_{L1}, g_{L2}, \ldots, g_{Ln}\}$, where each $g_{Li}$ corresponds to a label $Li$. And we denote the increase of $W_{in}^{Li}$ as $W_p$ when the local model is trained with the samples of $Li$. The decreases on the same $W_{in}^{Li}$ are $W_n$s when the local model is input with the samples of other labels, and both $W_p$ and $W_n$ are vectors too. But their magnitudes are different, i.e., the extent of increase is much higher than that of decrease. The specific values of weight update magnitudes may be changing in different training rounds. Nevertheless, the information about magnitudes in different training rounds can be obtained by training local models on the current global model, just like what the attacker does in Class Sniffing. 

As it happens, the positive effect of the increase can be offset by the accumulated impact of other decreases, and this phenomenon appears when a label is possessed by a small number of clients. However, we can still launch the following attack with the existence of above phenomenon. The details of such Quantity Inference attack is described in Algorithm~\ref{alg:2} and explained below.

The changes of inputting neuron weights do reflect the quantity information about training data, but not all of them possess such evident reflection, which means part of weights increase less than the rest and sometimes they could decrease even if the corresponding label appears in training. This set of weights are easily to experience the aforementioned `Offset' phenomenon, which could make the attack fail. Hence, the first question from the attacker perspective is how to remove them from the original intact inputting neuron weights. First, when we train the network with the data of a certain label in the training process, its existence will make the corresponding inputting neuron weights grow while the inputting neuron weights of other labels decrease. Following the simple superposition rules, the higher the ratio $R_{np} = |W_{n}| / |W_{p}|$ between magnitudes of $W_n$ and $W_p$ is, the easier `Offset' phenomenon emerges. Thus, we can set a threshold $Th_R$ towards $R_{np}$, and compare $W_p$ on each inputting weight $w_i$ in the weight vector $W_{in}$ with the $W_n$s on the same $w_i$ corresponding to other local updates. If there is an outlier whose corresponding ratio $R_{np}$ is higher than $Th_R$ , we will delete it from the original set $N_{int}^o$ and get a new set $N_{int}$, as shown in Algorithm~\ref{alg:2} from Line~\ref{de1_start} to Line~\ref{de1_end}.

\begin{figure}[!t]
\centering
\includegraphics[width=3.4in]{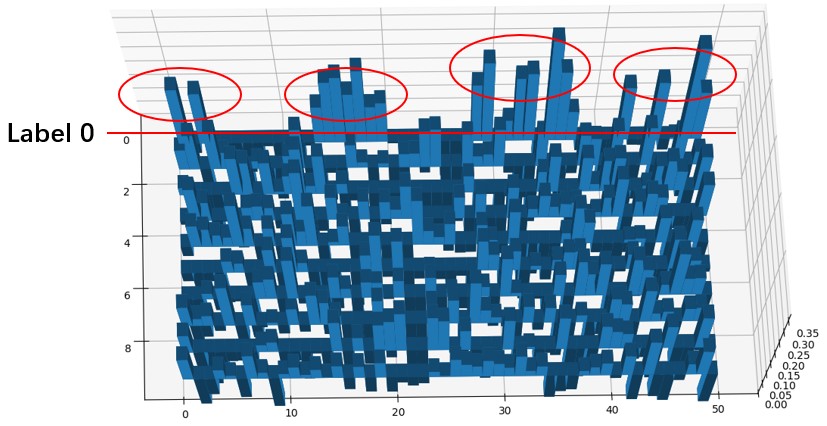}
\caption{\normalsize The inputting weight updates of the output layer (its former layer has 50 neurons) when the model is trained with the samples of label 0. And the updates circled by red ellipse are $N_{int}^{**}$.}
\label{fig_neurons}
\end{figure}

Next, let us take a label $L_{*}$ as an example. After local training process using auxiliary data, there will be following local updates $\{g_{L1}, \ldots, g_{L*}, \ldots, g_{Ln}\}$. Correspondingly, its original updates of the inputting weights are $N_{int}^{o*}$, which is a vector with the size equal to the number of neurons in the layer before the output. Then, we can regard the members of $N_{int}^{o*}$ on $g_{L*}$ as $W_p$ , and the updates vectors of the same $W_{in}^{L_*}$ on other $g_L$s as $W_n$s, followed by the process of deleting the aforementioned outliers and obtaining the new set $N_{int}^{*}$. Next, we are able to regard each member of $N_{int}^{*}$ on $g_{L*}$ as $w_p^{*}$, which denotes the increase when label $L_*$ is owned by a single client, and the averaged value of each member of $N_{int}^*$ on other $g_L$s as $w_n^*$, which indicates the negative impact of other labels. With $w_p^*$ and $w_n^*$, we can calculate all possible numbers of clients who own label $L_*$ by using each inputting weight change $N_i$ in $N_{int}^*$. The client number calculation formula is (\ref{formula_cal}), which is a derivation form of the simple average aggregation shown in (\ref{formula_aggregation}). 
\begin{equation}
w_p^* \cdot x + (N_p \cdot P \cdot N_L - x) \cdot w_n^* = N_p \cdot P \cdot G_{t+1} 
\label{formula_aggregation}
\end{equation}

\begin{equation}
x = \frac{N_p \cdot P \cdot (G_{t+1} - N_L \cdot w_n^*)}{w_p^* - w_n^*}    
\label{formula_cal}
\end{equation}
Here, $N_L$ indicates the average number of labels owned by selected clients, which is the same as that in Algorithm~\ref{alg:1}. $x$ is the predicted number of clients, and each $N_i$ corresponds to such a $x$. However, there are still abnormal weight changes whose corresponding $x$s are unreasonable. For instance, providing that there are $N_p \cdot P$ clients in a particular round, some $x$s could be larger than $N_p \cdot P$ or less than $0$ (the circumstance of less than $0$ is regarded as outlier since we assume label $L_*$ has been proven by Class Sniffing to be present in training). Thus, we also need to remove them from the current weight change set and obtain a final version $N_{int}^{**}$, which is shown in Figure~\ref{fig_neurons}, and the detailed steps can be seen in Algorithm~\ref{alg:2} from Line~\ref{de2_start} to Line~\ref{de2_end}. The final number of clients who owns label $L_*$ can be determined by the mean of $x$s corresponding to $N_{int}^{**}$. Another point worth mentioning is that the standard deviation of $x$s corresponding to $N_{int}^{**}$ should ideally be small, however occasionally it is large, 
in which case we abort the Quantity Inference attack. Such scenario happens at an extremely low frequency (below 1\% in the whole training process). 

\begin{algorithm}[ht]
\caption{Quantity Inference Attack}
\label{alg:2}
\begin{algorithmic}[1]
\REQUIRE ~~\\ 
Attacker's auxiliary data samples of different training labels, $D_{L0}, D_{L1},\ldots, D_{Ln}$;\\
Approximate number of labels owned by selected clients in a training round, $N_{L}$;\\
Selection proportion of clients in a training round, $P$;\\
Approximate number of whole participants, $N_{p}$;\\
Inputting weight positions of output layer each label, $W_{in}^{Li}$;\\
The interest label, $L_*$;\\
Ratio threshold $Th_R=0.5$;
\ENSURE ~~\\ 
The number of clients who own $L_*$, $X_*$;\\
\STATE \textbf{Begin:}
\STATE {Receive $\theta_t$ from server} 
\STATE $W$ = []
\FOR{$i$ in $D_{L0}, D_{L1},\ldots, D_{Ln}$}
\STATE $g_i$ = Local\_train($\theta_t$, $i$) - $\theta_t$
\STATE $w$ = acquire($g_i$, $W_{in}^{Li}$) $//$ acquire updates from $g_i$;
\STATE $W$.append($w$)
\ENDFOR
\STATE $N_{int}^*$ = []
\label{de1_start}
\STATE $x_*$ = []
\STATE $W_{temp}$ = delete($W$, $L^*$) $//$ delete the updates of $L^*$;
\FOR{each $w_i$ in $W_{in}^{L_*}$}
\STATE $w_p$ = $W$[$L^*$][$w_i$]
\STATE $w_n$ = mean($W_{temp}$)[$w_i$]
\STATE $R_{np}$ = $\mid w_n \mid$ / $\mid w_p \mid$
\IF{$R_{np} \leq Th_R$}
\STATE $N_{int}^*$.append($w_i$)
\ENDIF
\ENDFOR
\label{de1_end}
\FOR{each $N_i$ in $N_{int}^*$}
\label{de2_start}
\STATE $w_p^*$ = $W$[$L^*$][$N_i$]
\STATE $w_n^*$ = mean($W_{temp}$)[$N_i$]
\STATE $x_i = N_p \cdot P \cdot (G_{t+1} - N_L \cdot w_n^*) / (w_p^* - w_n^*)$
\IF{$x_i > N_p \cdot P$ or $x_i < 0$}
\STATE delete($N_{int}^*$, $N_i$) $//$ delete $N_i$ from $N_{int}^*$;
\ENDIF
\ENDFOR
\label{de2_end}
\STATE $N_{int}^{**}$ = $N_{int}^*$
\FOR{each $N_i$ in $N_{int}^{**}$}
\STATE $w_p^*$ = $W$[$L^*$][$N_i$]
\STATE $w_n^*$ = mean($W_{temp}$)[$N_i$]
\STATE $x_i = N_p \cdot P \cdot (G_{t+1} - N_L \cdot w_n^*) / (w_p^* - w_n^*)$
\STATE $x_*$.append($x_i$)
\ENDFOR
\STATE $X_*$ = mean($x_*$)
\end{algorithmic}
\end{algorithm}

\subsection{Whole Determination}
If the attacker is not sensitive about time immediacy and patient enough, which means he cares about the composition proportion of entire training data over a long training span rather than just a single or several training rounds, we can propose another new attack. This attack lays emphasis on the overfitting characteristic of learning model when the training process sustains constantly, which suits best to FL application scenarios.

\begin{figure}[!t]
\centering
\includegraphics[width=3.4in]{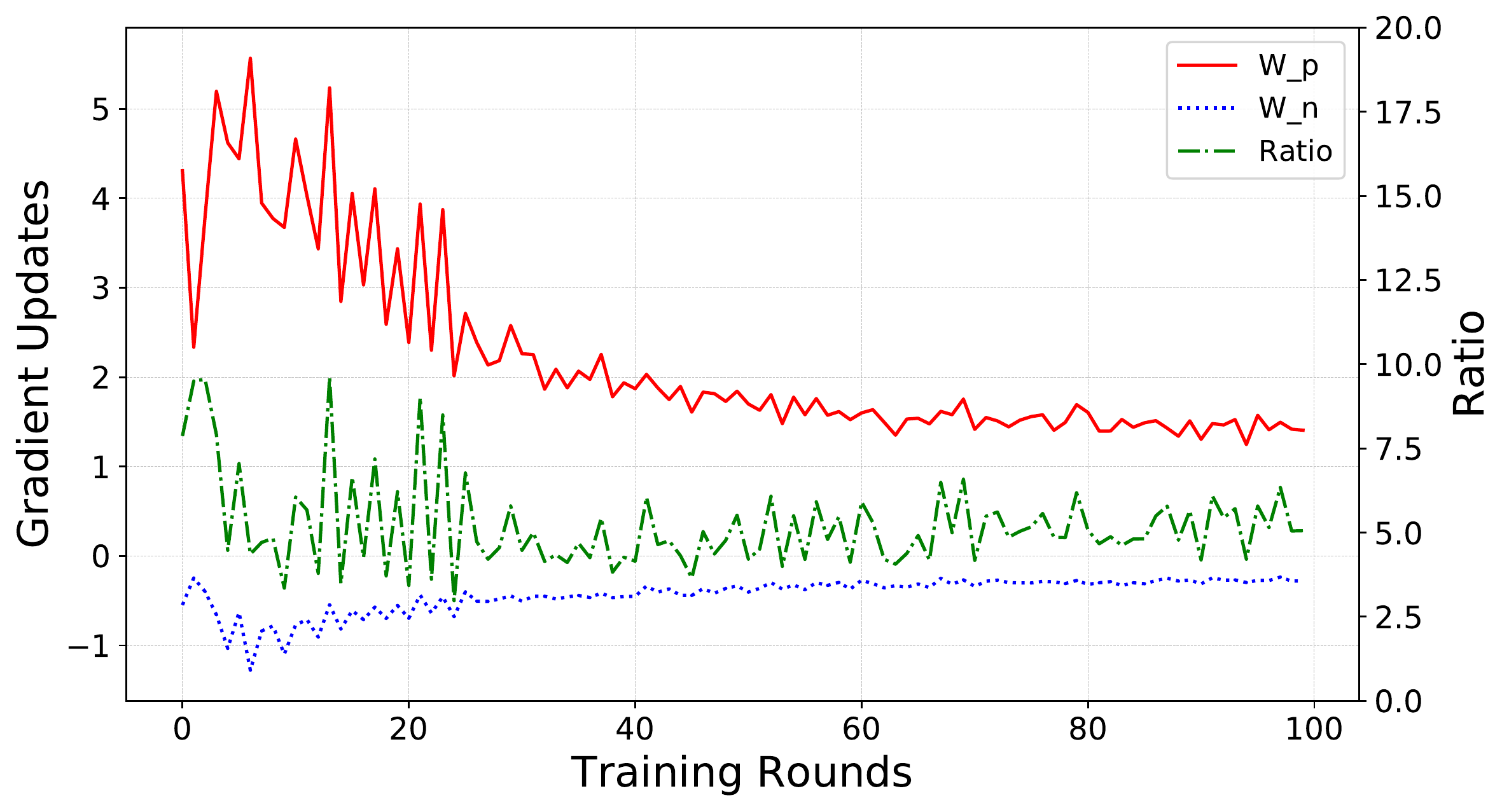}
\caption{\normalsize The ratio $(|W_p| / |W_n|)$ change of a frequent label among training rounds.}
\label{fig_frequent_label}
\end{figure}

\begin{figure}[!t]
\centering
\includegraphics[width=\linewidth]{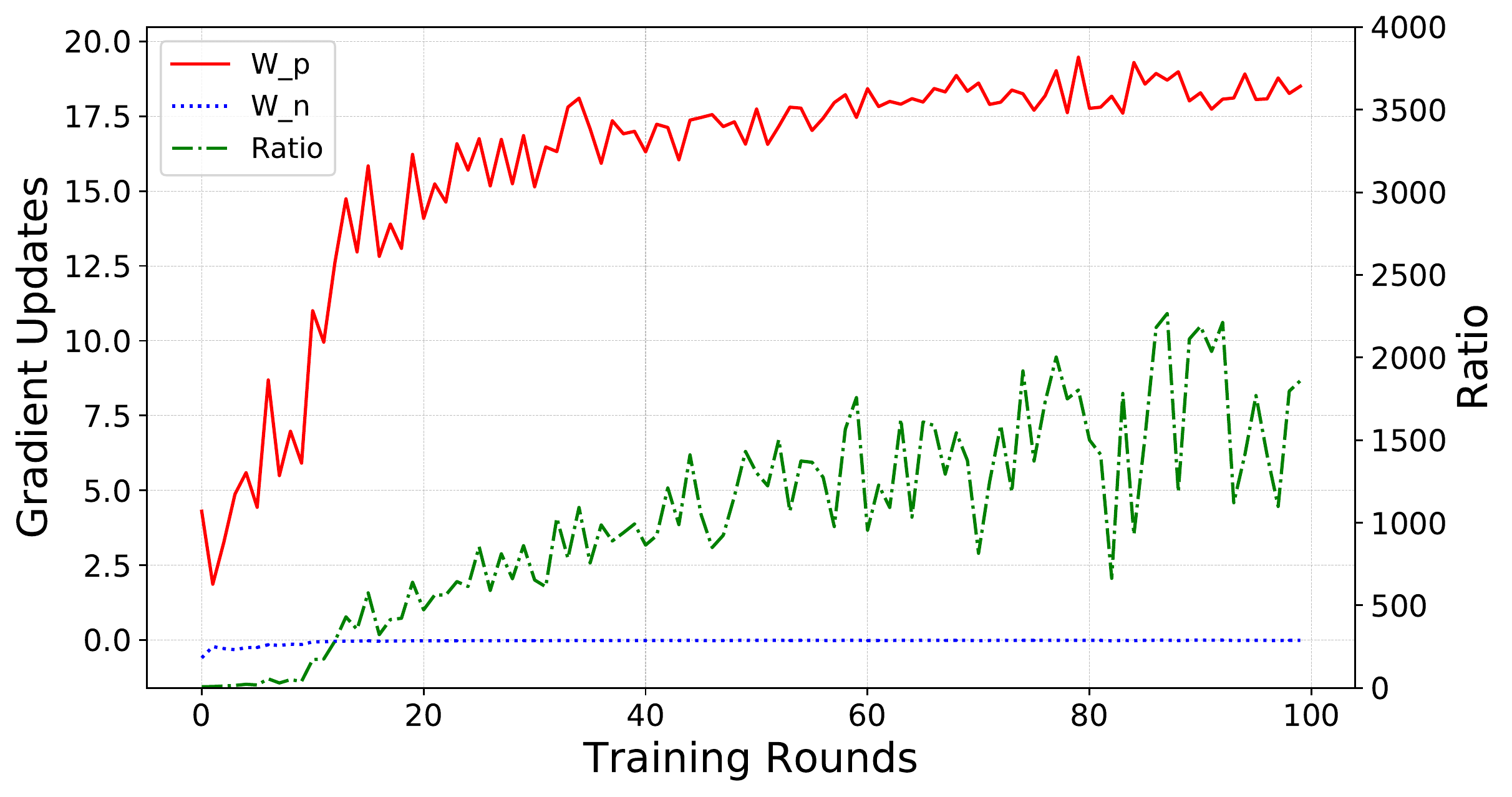}
\vspace{-15pt}
\caption{\normalsize The ratio $(|W_p| / |W_n|)$ change of an occasional label among training rounds.}
\label{fig_occasional_label}
\end{figure}

Let us describe an example case. In the training process of a deep neural network, there are a set of labels appearing frequently (the number of samples is large) and other labels appearing occasionally (the number of samples is small), denoted by $\{L_{f1}, L_{f2}, \ldots, L_{fk1}\}$ and $\{L_{o1}, L_{o2}, \ldots, L_{ok2}\}$, respectively. Like the former attacks, the attacker downloads current global model $G_t$ and trains his local one with auxiliary data inputted label by label, and eventually he can obtain all corresponding local gradient updates of each label, i.e., $\{g_{f1}, g_{f2}, \ldots, g_{fk1}, g_{o1}, g_{o2}, \ldots, g_{ok2}\}$. When we investigated the inputting weight changes of one particular frequent label and an occasional label, for instance, $N_{int}^{f1}$ of $L_{f1}$ and $N_{int}^{o1}$ of $L_{o1}$, we observed an interesting phenomenon that the corresponding absolute value of $N_{int}^{f1}$ and $N_{int}^{o1}$ on other $g$s (except $g_{f1}$ and $g_{o1}$) present a huge difference. That is to say that the absolute values of $N_{int}^{f1}$ in other frequent gradient updates, i.e., $\{g_{f2}, \ldots\}$, are much higher than those of $N_{int}^{o1}$ in the same gradient updates $\{g_{f2}, \ldots\}$. This is from the perspective of $W_n$. Similar huge difference can be seen for $W_p$ (e.g., the difference between the $N_{int}^{f1}$ on $g_{f1}$ and the $N_{int}^{o1}$ on $g_{o1}$). In various experiments, such phenomenon can be easily observed. For instance, it appears after 10 epochs in the MNIST classifier training process, which is shown in Figure~\ref{fig_frequent_label} and Figure~\ref{fig_occasional_label}. The attacker is then able to analyze this phenomenon to access the information about the composition proportion of training labels. 

\subsubsection{Explanation}
Let us give a possible explanation about this phenomenon here. The connection between inputting weight updates of the output layer and the corresponding labels is strong enough that we can regard the neuron weights $x_{in}$ as the features set of each label. Then, (\ref{formula_y}) is changed to \begin{equation}
    Y = (f_{f1}(x_{in}^{f1}), ..., f_{fk1}(x_{in}^{fk1}), f_{o1}(x_{in}^{o1}), ..., f_{ok2}(x_{in}^{ok2}))\label{formula_4}
\end{equation}
\begin{equation}
    F = (g_{f1}(x_{in}^{f1}), ..., g_{fk1}(x_{in}^{fk1}), g_{o1}(x_{in}^{o1}), ..., g_{ok2}(x_{in}^{ok2}))
\end{equation}
Here $k1 + k2 = k$. 
We assume that the features embedded in the neurons of the output layer are highly independent of each other after the extract and filter of front layers. This independence is regarded as they are irrelevant to each other, hence their derivatives to each other are zero:
\begin{equation}
    \frac{\partial x_{in}^m}{\partial x_{in}^n} = 0, (m \neq n)
\end{equation}
The output format of a classifier is a type of probability vector, and the outputting result of a particular input is the label corresponding to the highest dimension in the probability vector. Then, we know that the quantities of data samples in terms of different labels are different, hence their proportions in the target label $y$ are different. We thus hypothesize that the proportion can be measured by calculating the derivative of $y$ on the features of the output layer. If a label has more samples, its proportion will be greater, and vice versa. Then, it is presented as 
\begin{equation}
    \frac{\partial Y}{\partial x_{in}^{fi}} \gg \frac{\partial Y}{\partial x_{in}^{oj}}, (i \in (0, k1)\ \&\ j \in (0, k2))
\end{equation}
Then the specific format of the updates is $|\nabla_{f1} Loss(\theta)^{f2}|$ and $|\nabla_{o1} Loss(\theta)^{f2}|$, which corresponds to the aforementioned phenomenon from the perspective of $W_n$.
\begin{equation*}
\begin{aligned}
    \left|\nabla_{f1} Loss(\theta)^{f2}\right| &= \frac{\partial Loss(\theta)^{f2}}{\partial x_{in}^{f1}} + \sum_{i=2}^{k1} \frac{\partial Loss(\theta)^{f2}}{\partial x_{in}^{fi}} \cdot \frac{\partial x_{in}^{fi}}{\partial x_{in}^{f1}}\\
    & + \sum_{j=1}^{k2} \frac{\partial Loss(\theta)^{f2}}{\partial x_{in}^{oj}} \cdot \frac{\partial x_{in}^{oj}}{\partial x_{in}^{f1}} = \frac{\partial D(F, Y_{f2})}{\partial x_{in}^{f1}}\\
    & = \frac{\partial D(Y_{f2})}{\partial F} \cdot \frac{\partial F}{\partial x_{in}^{f1}} + \frac{\partial D(F)}{\partial Y_{f2}} \cdot \frac{\partial Y_{f2}}{\partial x_{in}^{f1}}\\
    & = \frac{\partial D(Y_{f2})}{\partial F} \cdot \frac{\partial g_1(x_{in}^{f1})}{\partial x_{in}^{f1}} \\
    & + \frac{\partial D(F)}{\partial Y_{f2}} \cdot \frac{\partial f_1^{f2}(x_{in}^{f1})}{\partial x_{in}^{f1}}
\end{aligned}
\end{equation*}

\begin{equation*}
\begin{aligned}
    |\nabla_{o1} Loss(\theta)^{f2}| &= \frac{\partial D(Y_{f2})}{\partial F} \cdot \frac{\partial F}{\partial x_{in}^{o1}} + \frac{\partial D(F)}{\partial Y_{f2}} \cdot \frac{\partial Y_{f2}}{\partial x_{in}^{o1}}\\
    & = \frac{\partial D(Y_{f2})}{\partial F} \cdot \frac{\partial g_1(x_{in}^{o1})}{\partial x_{in}^{o1}} \\
    & + \frac{\partial D(F)}{\partial Y_{f2}} \cdot \frac{\partial f_1^{f2}(x_{in}^{o1})}{\partial x_{in}^{o1}}
\end{aligned}
\end{equation*}

The $Y_{f2}$ here denotes the next version of ideal model when the current global model accepts the input of label $L_{f2}$ training samples, hence it is natural that there is only difference existing in the derivatives of $f_2(x_{in}^{f2})$ between these two versions to some extent. In other words, the differences of other $f$s can be neglected:
\begin{equation}
\begin{aligned}
    f_1^{f2}(x_{in}^{f1}) = f_1(x_{in}^{f1})\\
    f_1^{f2}(x_{in}^{o1}) = f_1(x_{in}^{o1})
\end{aligned}
\end{equation}
Since $L_{f1}$ and $L_{f2}$ are frequent labels, $L_{o1}$ is an occasional label, and $F$ is the current global model, which is the same in both cases, we can obtain that

\begin{equation}
    \frac{\partial g_1(x_{in}^{f1})}{\partial x_{in}^{f1}} \gg \frac{\partial g_1(x_{in}^{o1})}{\partial x_{in}^{o1}}
\end{equation}

\begin{equation}
    \frac{\partial f_1^{f2}(x_{in}^{f1})}{\partial x_{in}^{f1}} = \frac{\partial f_1(x_{in}^{f1})}{\partial x_{in}^{f1}} \gg \frac{\partial f_1^{f2}(x_{in}^{o1})}{\partial x_{in}^{o1}} = \frac{\partial f_1(x_{in}^{o1})}{\partial x_{in}^{o1}}
\end{equation}

Combine the above results, we can conclude that 
\begin{equation}
\label{large}
    |\nabla_{f1} Loss(\theta)^{f2}| \gg |\nabla_{o1} Loss(\theta)^{f2}|
\end{equation}

\subsubsection{Attack Approach}
Based on the phenomenon above (\ref{large}), we can conclude that the ratio $R_{pn}=|W_{p}| / |W_{n}|$ of occasional labels is larger than that of frequent labels, and leverage such conclusion to determine the quantity relation between different labels. Like former attacks, when the adversary launches the Whole Determination attack, he trains his local model on the basis of $G_t$ with auxiliary data $D_{L0}, D_{L1},\ldots, D_{Ln}$, and correspondingly obtains local updates $\{g_{L1}, g_{L2}, \ldots, g_{Ln}\}$. Then, he can calculate all $R_{pn}$s by using data from $\{g_{L1}, g_{L2}, \ldots, g_{Ln}\}$. Next, these $R_{pn}$s will be formed as a vector $V_{pn}$, where each label corresponds to a vector $V_{pn}^i$. Finally, these vectors can be clustered into different groups by an unsupervised algorithm, and the vectors being in the same group indicates their corresponding labels have approximately the same number of data samples in training. The quantity could present huge differences if labels belong to different clusters. 
Here, the unsupervised algorithm we adopted is Hierarchical Clustering, which can classify given data into different clusters with the metric of Euclidean Distance. Attackers may also choose other clustering approaches.

\section{Evaluation}
All experiments were conducted on a workstation running Ubuntu 18.04 LTS equipped with a 2.10GHz CPU Intel Xeon(R) Gold 6130, 64GB RAM, and an NVIDIA TITAN RTX GPU card. We construct the model mainly on PyTorch \cite{paszke2017pytorch}, and use Scipy-scikit-learn \cite{pedregosa2011scikit} to implement some machine learning models.

\subsection{Experiment Setting}
\subsubsection{Auxiliary Data}
Like many other inference attacks on machine learning, the attacker also needs auxiliary data. It consists of some data samples of the labels he wants to infer. In practice, such data samples are often not difficult to acquire. 
The number of data samples should be close to the average quantity owned by participants. If the samples are not enough, the attacker may try reproduction techniques such as GAN to construct more similar samples.

\subsubsection{Network Structure}
The main structure is based on standard construction of federated learning~\cite{konevcny2016federated}, with some modifications for practical purpose, e.g., participating clients are able to process several epochs locally rather than just a single epoch before sending their updates to the aggregation server~\cite{mcmahan2016communication}. The symbols of major hyper-parameters are defined in Table~\ref{table_1}.
\begin{table}[!t]
\renewcommand{\arraystretch}{1.5}
\caption{Hyper-parameters of FL in evaluation}
\label{table_1}
\centering
\begin{tabular}{cl}
\hline
\textbf{Symbols} & \textbf{Description}\\
\hline
$L_{bs}$ & Local model training batch size\\
\hline
$L_{lr}$ & Local model learning rate\\
\hline
$L_{ep}$ & Local model training epochs\\
\hline
$P$ & Selection proportion of clients in a training round\\
\hline
$model$ & Selected models to accomplish learning task\\
\hline
$N_p$ & Approximate number of whole participants\\
\hline
\end{tabular}
\end{table}
\subsubsection{Datasets}
\label{dataset}
The dataset information (number of training labels and corresponding training model) are presented in Table~\ref{table_2}. We choose these datasets that are close to our concerns about privacy in daily life. For instance, Fer2013 is relevant to face recognition, while HAM10000 aims at diagnosing several skin cancers. Both of them contain private information owned by different people. 
\begin{table}[!t]
\renewcommand{\arraystretch}{1.5}
\caption{Datasets and relevant information in our experiments}
\label{table_2}
\centering
\begin{tabular}{cccl}
\hline
\textbf{Datasets} & \textbf{\#Records} & \textbf{\#Labels} & \textbf{Model}\\
\hline
\textbf{MNIST} & $60.0K$ & $10$ & \textbf{MLP \& CNN}\\
\hline
\textbf{CIFAR10} & $60.0K$ & $10$ & \textbf{LeNet5 \& Resnet18}\\
\hline
\textbf{Fer2013} & $28.7K$ & $7$ & \textbf{Resnet18}\\
\hline
\textbf{HAM10000} & $37.0K$ & $7$ & \textbf{Resnet18}\\
\hline
\end{tabular}
\end{table}

\begin{itemize}

\item \textbf{MNIST.} As one of the most popular and classical datasets in machine learning, \label{default} MNIST includes 10 labels, each of which corresponds to approximately 6,000 32$\times$32 gray handwritten digital images. 5,000 of them are training data, while 1,000 are for testing. Because of its simplicity and the small number of total training data, it is not easy for deep and complicated models to achieve high performance. Hence we choose a standard but simple MLP (multi-layer perceptron) and a CNN model for it, both of which are able to achieve $98\%$ accuracy.

The MLP model contains an input layer, followed by two fully-connected hidden layers of size 256 and 64. There is a dropout operation between hidden layers, and finally it has an output layer with size of 64. We use rectified linear unit (ReLU) as the activation function for all layers. Other settings are $L_{bs}=64$, $L_{lr}=0.01$, $L_{ep}=3$. The CNN model consists of two spatial convolution layers with 10 and 20 filters (kernel size is 5$\times$5), max pooling layers with size set to 2, a dropout layer and a fully-connected layer with size 320, and finally an output layer whose size is 50. The activation function is ReLU, and other settings are the same as MLP.

\item \textbf{CIFAR10.} CIFAR10 consists of 10 classes containing 6,000 32$\times$32 RGB images each, which can also be divided into 5,000 for training and 1,000 for testing. The entire training labels contain common objects in daily life, suitable for the object identifying task on smartphones. For clustering model, we select two commonly used networks, LeNet5~\cite{lecun1998gradient} and ResNet18. The former can achieve $85\%$ accuracy, while the latter can achieve $98\%$.
LeNet5 consists of two convolution layers with 6 and 16 filters respectively (kernel size of them is 5$\times$5), pooling layers with size set to 2, two fully-connected linear layers with size set to 400 and 120, and an output layer and a softmax layer. Its parameter setting is $L_{bs}=64$, $L_{lr}=0.1$, $L_{ep}=3$. The specific network structure of Resnet18 can be found in~\cite{he2016deep}. The parameter setting on Resnet18 is $L_{bs}=64$, $L_{lr}=0.1$, $L_{ep}=3$.

\item \textbf{Fer2013.} Fer2013 \cite{goodfellow2013challenges} originates from a Kaggle competition, which is Facial Expression Recognition Challenge 2013, and it aims to build a learning model to recognize human's expression automatically. It contains approximately 30,000 facial RGB images of different expressions with size restricted to 48$\times$48, and the main labels of it can be divided into 7 types: 0=Angry, 1=Disgust, 2=Fear, 3=Happy, 4=Sad, 5=Surprise, 6=Neutral. The Disgust expression has the minimal number of images -- 600, while other labels have nearly 5,000 samples each. We randomly select $80\%$ of the samples for each label as the training data, and use the rest for testing. We choose Resnet18 as the learning model for Fer2013. It is trained under the setting $L_{bs}=64$, $L_{lr}=0.02$, $L_{ep}=3$. The testing accuracy is $95\%$.

\item \textbf{HAM10000.} HAM10000 \cite{tschandl2018ham10000} is a large collection of multi-source dermatoscopic images of pigmented lesions. There are nearly 37,000 records about skin lesions, and they are classified into 7 labels, i.e., 0=Melanocytic nevi, 1=Melanoma, 2=Benign keratosis-like lesions, 3=Basal cell carcinoma, 4=Actinic keratoses, 5=Vascular lesions, 6=Dermatofibroma. Each label corresponds to approximately 5,000 images. Similarly, we randomly divide the data into 4,500 for training and 500 for testing. HAM10000 is also trained on Resnet18, and its parameters are $L_{bs}=128$, $L_{lr}=0.1$, $L_{ep}=3$. The testing accuracy is $90\%$.
\end{itemize}
\subsection{Class Sniffing}
In order to simulate more practical application scenarios of FL, we try to allocate training dataset samples randomly. Let us take MNIST as an example. We create a setting with 100 participants. In each training round, the server is required to select 10 clients randomly and collect their gradient updates to form an aggregated global model. In our setting, each participant can possess 3, 4 or 5 main labels and a small number of other labels, and the number of data samples per main label is much larger than that of other labels. All participants select their own main labels randomly. The data allocations of other datasets are similar to that of MNIST, which we believe can simulate the practical scenarios to some extent.

\begin{table*}[htbp]
\caption{The success rate of Class Sniffing on experiment datasets}
\label{table_3}
\centering
\begin{tabular}{|c|c|c|c|c|c|c|c|c|c|c|c|}
\hline
Dataset                  & Model    & \multicolumn{10}{c|}{Success Rate(\%)}               \\ \hline
\multicolumn{2}{|c|}{Labels}        & 0  & 1  & 2   & 3   & 4   & 5  & 6  & 7   & 8   & 9  \\ \hline
\multirow{2}{*}{MNIST}   & MLP      & 94 & 97 & 95  & 98  & 99  & 93 & 94 & 96  & 97  & 95 \\ \cline{2-12} 
                         & CNN      & 96 & 97 & 98  & 93  & 96  & 95 & 95 & 96  & 98  & 94 \\ \hline
\multirow{2}{*}{CIFAR10} & LeNet5      & 92 & 94 & 97  & 99  & 98  & 93 & 96 & 98  & 99  & 96 \\ \cline{2-12} 
                         & Resnet18 & 93 & 97 & 97  & 94  & 97  & 95 & 96 & 93  & 98  & 97 \\ \hline
Fer2013                  & Resnet18 & 99 & 94 & 95  & 94  & 97  & 98 & 98 & -   & -   & -  \\ \hline
HAM10000                 & Resnet18 & 93 & 97 & 98  & 98  & 98  & 96 & 95 & -   & -   & -  \\ \hline
\end{tabular}
\end{table*}

\begin{table*}[htbp]
\caption{The success rate of Quantity Inference on experiment datasets}
\label{table_4}
\centering
\begin{tabular}{|c|c|c|c|c|c|c|c|c|c|c|c|}
\hline
Dataset                  & Model    & \multicolumn{10}{c|}{Success Rate(\%)}               \\ \hline
\multicolumn{2}{|c|}{Labels}        & 0  & 1  & 2   & 3   & 4   & 5  & 6  & 7   & 8   & 9  \\ \hline
\multirow{2}{*}{MNIST}   & MLP      & 91 & 91 & 94  & 93  & 93  & 94 & 92 & 97  & 96  & 92 \\ \cline{2-12} 
                         & CNN      & 95 & 96 & 98  & 98  & 100 & 95 & 95 & 100 & 100 & 96 \\ \hline
\multirow{2}{*}{CIFAR10} & LeNet5      & 96 & 94 & 98  & 96  & 96  & 92 & 94 & 97  & 94  & 94 \\ \cline{2-12} 
                         & Resnet18 & 92 & 95 & 94  & 93  & 95  & 92 & 98 & 93  & 95  & 96 \\ \hline
Fer2013                  & Resnet18 & 97 & 92 & 100 & 100 & 98  & 95 & 98 & -   & -   & -  \\ \hline
HAM10000                 & Resnet18 & 95 & 93 & 94  & 95  & 93  & 94 & 97 & -   & -   & -  \\ \hline
\end{tabular}
\end{table*}

The goal of Class Sniffing is to predict whether a certain label appears in a training round, hence the evaluated metric here is the success rate of prediction. That is, if we correctly detect the existence of a label for $T_s$ times and fail for $T_f$ times in several training rounds, then the success rate is $T_s/(T_s+T_f)$. We perform this attack on all datasets with each own standard model for 100 training rounds, and the results are presented in Table~\ref{table_3}. As shown in the table, the success rate is relatively high (above $90\%$) for all datasets, which demonstrates the effectiveness of Class Sniffing.

\subsection{Quantity Inference}
The Class Sniffing attack is designed to detect the existence of a particular label, while Quantity Inference aims to acquire the quantity information of a label in a single training cycle. Because the pre-settings of Class Sniffing are relatively practical, there is no need to change them here. That is, the participants of FL are also set to 100, the randomly selected fraction is $10\%$, and the allocation strategy in terms of dataset samples is the same.
\subsubsection{Metrics}
\label{metric}
Considering both the threat model and the problems we want to solve, we need to define a new metric to evaluate the attack here. The main idea is to set an error bound and evaluate how often the attacker can estimate the number of clients possessing a particular label within the error bound. 
Specifically, in a particular training round, assume there are $i$ clients possessing a label $L$, and the attacker launches Quantity Inference in this round and obtains an estimated number $\hat{i}$ of clients who possess label $L$. We regard an attack successful if $|\hat{i} - i| \leq \alpha$, while failed when $|\hat{i} - i| > \alpha$, where the error bound $\alpha$ controls the accuracy requirement. We set $\alpha = 1$ in our experimental evaluation. Then, by recording the number of times the attacker successfully make the estimation (i.e., within the error bound) and fails it (i.e., larger than the error bound), we can calculate the success rate.
\subsubsection{Results}
We evaluates the Quantity Inference attack on all datasets with each own standard models for 100 training rounds, and the results are shown in Table~\ref{table_4}. We can see that the success rate is high for all datasets, i.e., between $91\%$ and $100\%$, which shows the effective of our Quantity Inference attack. Moreover, the results are consistently high across the four datasets, which shows the broad applicability of our approach. 

\subsubsection{Impact of Hyper Parameters}
\begin{figure}[!t]
\centering
\includegraphics[width=3.2in]{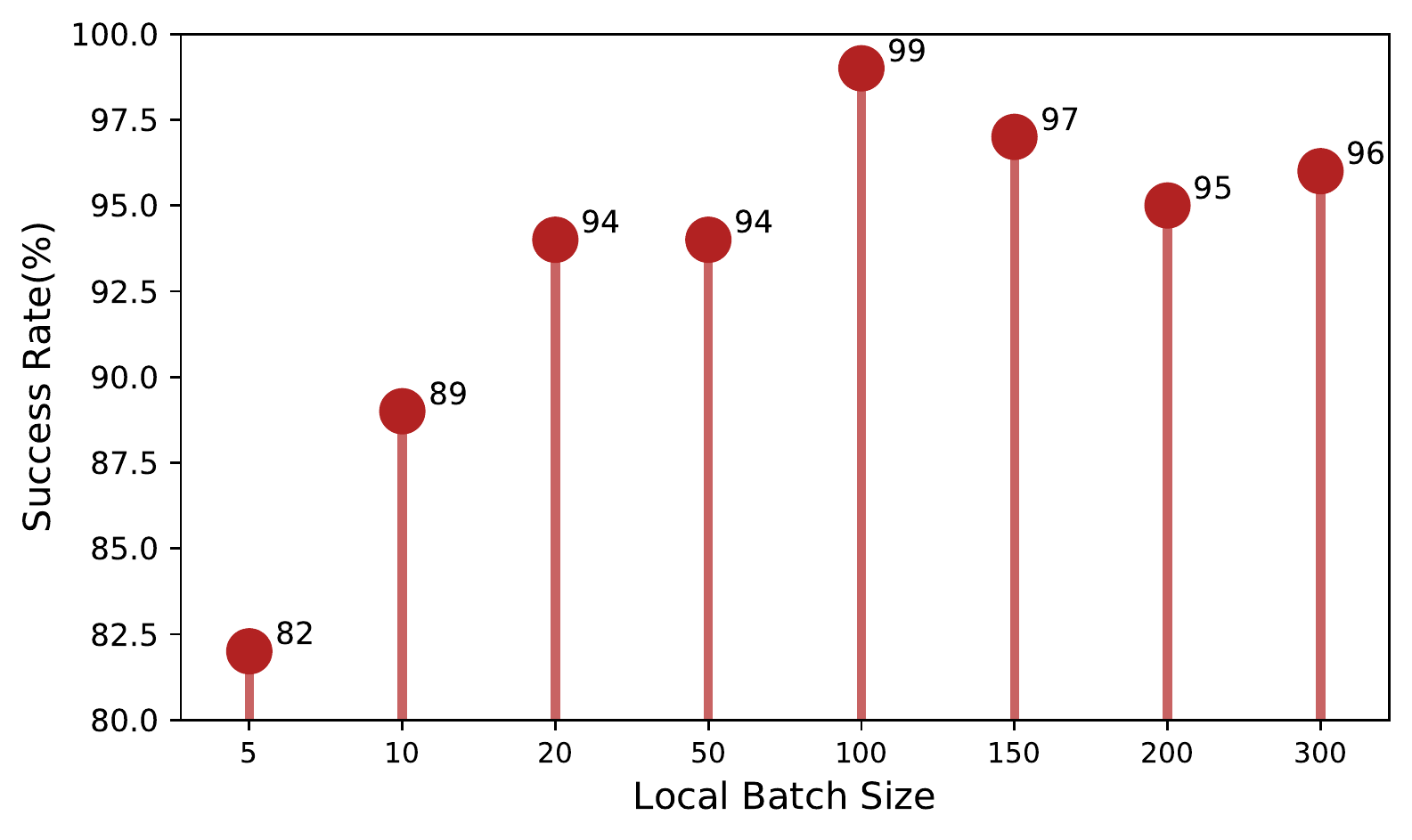}\vspace{-10pt}
\caption{\normalsize The success rate of Quantity Inference with different training batch sizes among 100 training rounds.}
\label{fig_batch_size}
\end{figure}

\begin{figure}[!t]
\centering
\includegraphics[width=3.2in]{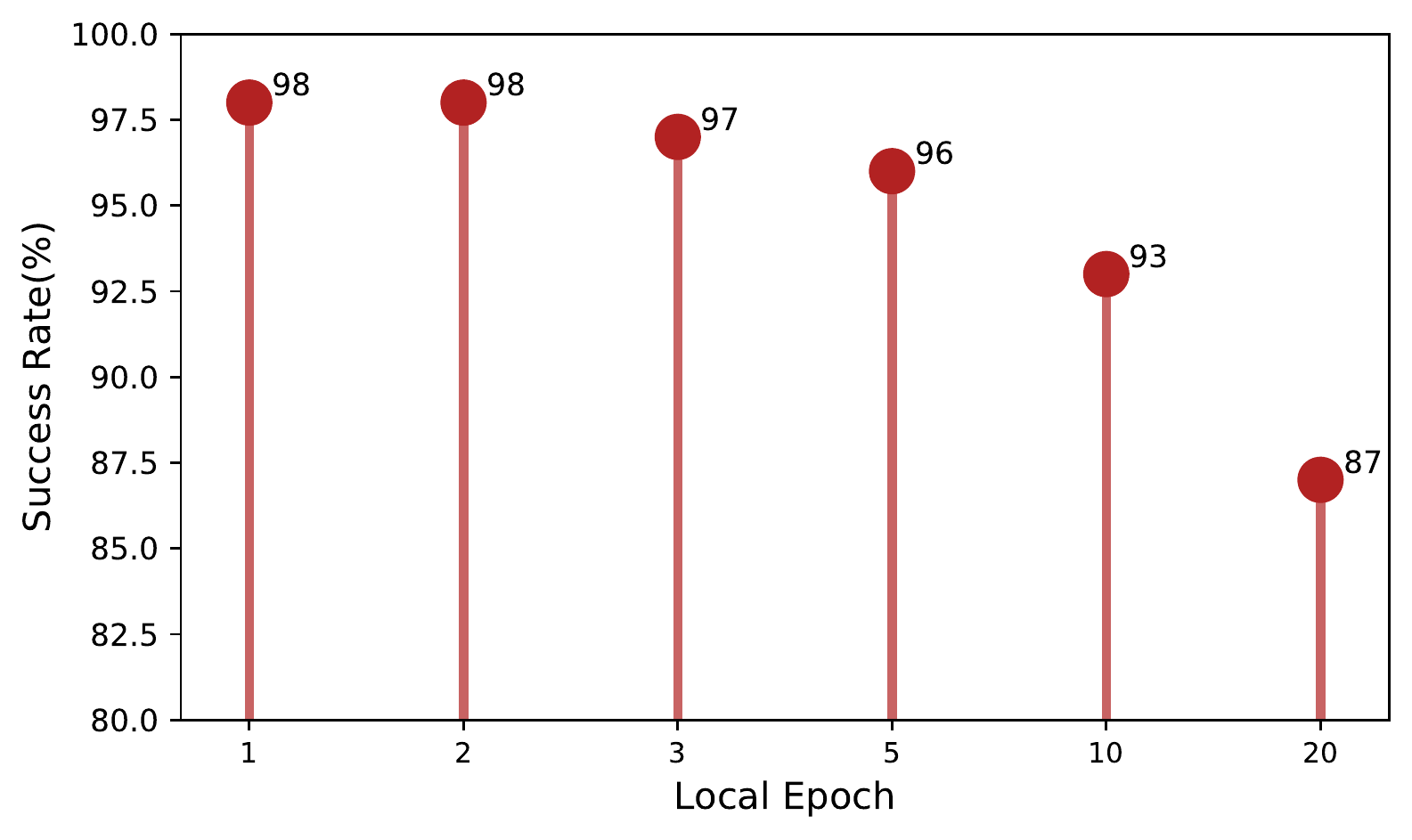}\vspace{-10pt}
\caption{\normalsize The success rate of Quantity Inference with different local training epochs among 100 training rounds.}
\label{fig_local_epoch}
\end{figure}

We also study how Quantity Inference is affected by hyper-parameters, in particular $L_{bs}$ (local model training batch size) and $L_{ep}$ (local model training epochs). We choose MNIST with CNN for our study, and fix other settings as in Sec.~\ref{default} when we evaluate a particular parameter. The results are shown as Figure~\ref{fig_batch_size} and Figure~\ref{fig_local_epoch}. 

For the impact of different batch sizes shown in Figure~\ref{fig_batch_size}, we can observe that success rate hits the bottom ($82\%$) when the batch size is set to relatively small, and reaches a relatively high level ($94\%$) with batch size at 20. As we know, the batch size in training should not be set too small as it may lead to more calculation iterations and cause the model perform bad. Hence, we think Class Sniffing should be effective under common batch size settings.

We can also clearly observe the impact of local training epochs in Figure~\ref{fig_local_epoch}, where the success rate decreases slightly when the local epoch rises (the lowest is $87\%$). To our knowledge, the standard application of FL usually allows participants to train their local model only one epoch. The reason is that considering the limited computation capacity of local devices, more epochs will drastically increase the computation load and may affect the normal operation of those devices. 
Recently, some researchers propose that local devices can shoulder more computation tasks~\cite{mcmahan2016communication}, which means more local training epochs are possible, however more than 10 epochs would still be quite demanding. Thus, we believe that Quantity Inference should work well in most circumstances, even when there are multiple local training epochs (as long as not too many).

\subsubsection{Quantity of Participants}
\begin{figure}[!t]
\centering
\includegraphics[width=3.2in]{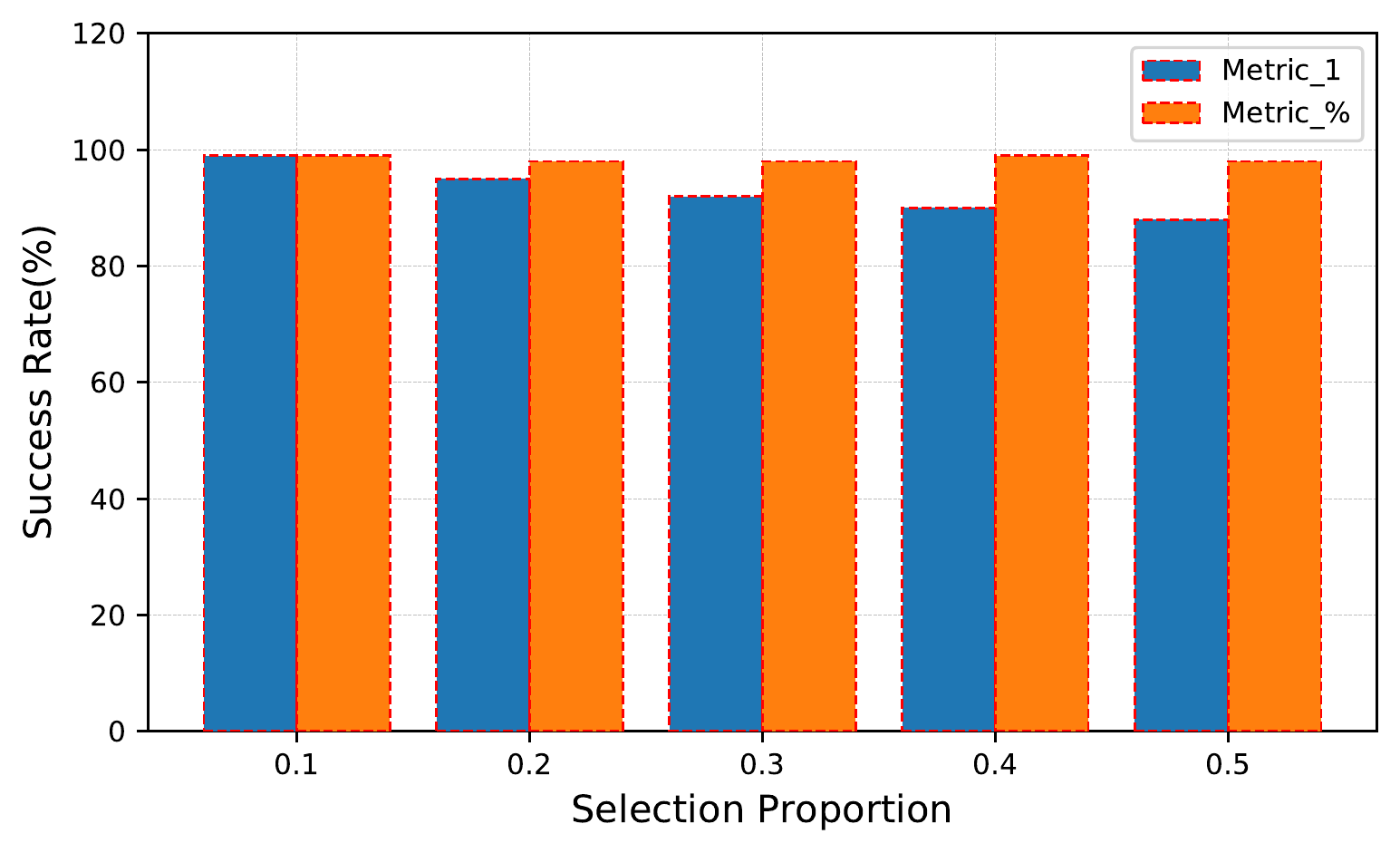}
\caption{\normalsize The success rate of Quantity Inference with different selection proportions among 100 participants.}
\label{fig_qi_frac}
\end{figure}

\begin{figure}[!t]
\centering
\includegraphics[width=3.2in]{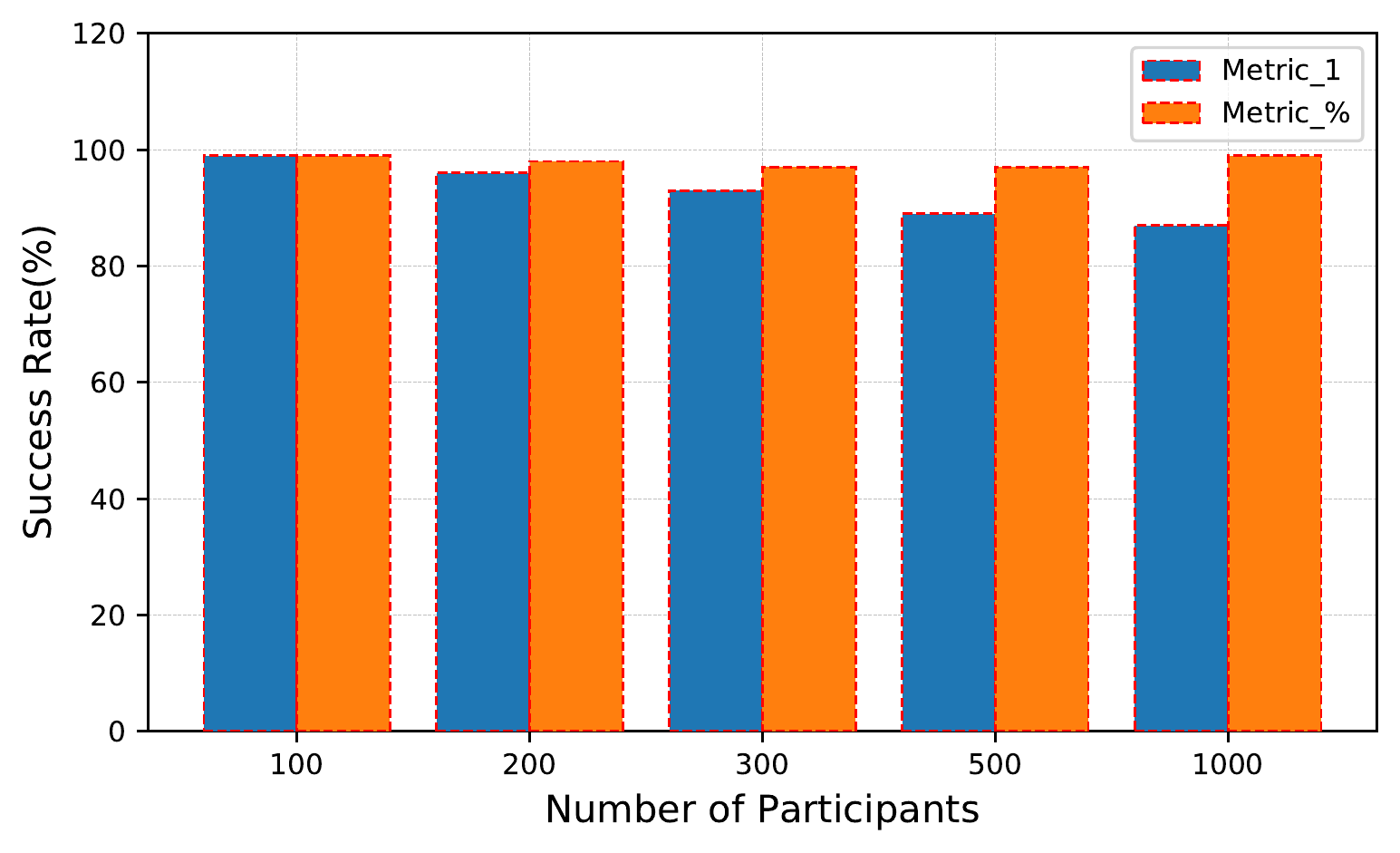}
\caption{\normalsize The success rate of Quantity Inference with different numbers of participants with selection proportion $10\%$.}
\label{fig_qi_num}
\end{figure}

To further investigate the practicality of Quantity Inference, we consider changing the selection proportion $P$ and the overall number of participants $N_p$. In our former default settings, $P = 10\%$ and $N_p = 100$. In this study, we explore a range of values, and the results are shown in Figure~\ref{fig_qi_frac} and Figure~\ref{fig_qi_num}.

We first fix the overall number of participants to 100, and change the selection fraction from 0.1 to 0.5 with the step of 0.1. If we still use the original metric in Sec.~\ref{metric}, which is denoted as $Metric\_1$ in the figures, we can see that the success rate shows a moderate decline from $99\%$ to $88\%$ when the proportion is increasing. Such trend is reasonable, as the more clients there are in each round, the harder it is to achieve the same success rate under an absolute error bound. However, we could consider a new metric $Metric\_\%$ based on a relative error bound (depending on the number of participants), which is defined as the difference between the real number of clients possessing a label and the estimated number with respect to an error bound $\alpha=5\% \cdot N_p \cdot P$ rather than $\alpha = 1$ of $Metric\_1$. Under $Metric\_\%$, the success rate always stays at a high level (near $100\%$).

Then, we make the selection proportion unchanged at $10\%$, and increase the number of overall participants from 100 to 1000. We can observe the trend that the success rate slightly decreases when the number of participants increases. However, if we use the metric $Metric\_\%$ based on relative error bound, we can see that the success rate stays at a high level (near $100\%$). Overall, both figures demonstrate the effectiveness of our Quantity Inference attack over a wide range of number of overall participants and selection proportion.


\subsection{Whole Determination}
Whole Determination is an attack typically towards developed models, which have been trained with considerable data samples and perform well on corresponding given tasks. Hence, for its evaluation, we choose to launch the attack in the middle and late stages of the training process when the model is near to convergence. However, it does not mean that Whole Determination cannot work in more advanced stages. Before the attack, all datasets above are required to train themselves under their own default settings. When the loss of model decreases to a relative small value, the attacker will use Whole Determination to obtain the composition proportion of training labels. 

\subsubsection{Dataset Allocation}
\begin{figure}[!t]
\centering
\includegraphics[width=3.2in]{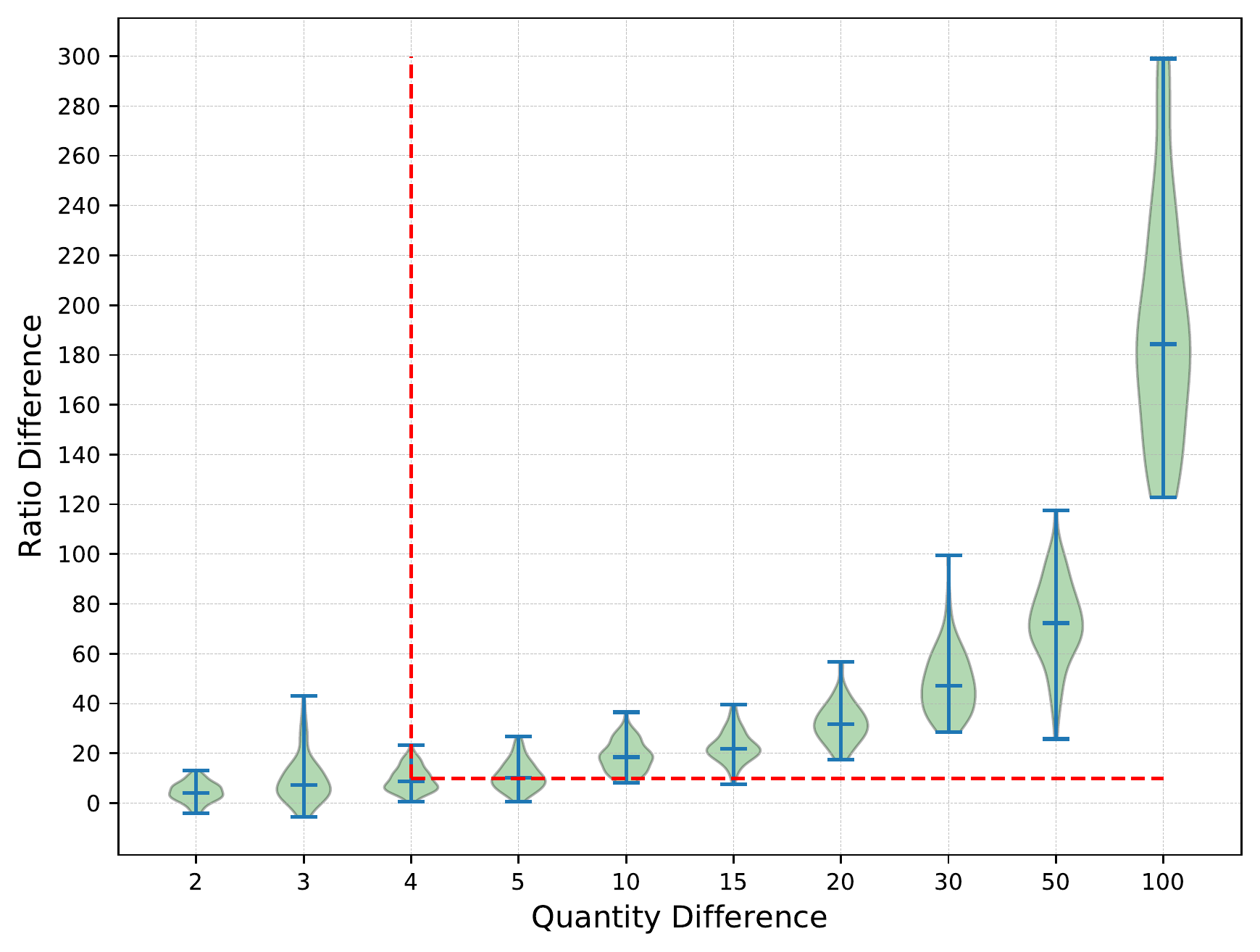}
\caption{\normalsize The ratio difference between two labels, e.g., label $L1$ and $L2$, and the difference is $Ratio1 - Ratio2$. The index of horizontal axis ($2, 3, 4, 5, ..., 100$) means the number of data samples owned by $L1$ is $\{1/2, 1/3, 1/4, 1/5, ..., 1/100\}$ of that owned by $L2$.}
\label{fig_wd_ratio}
\end{figure}

In previous experiments, the number of data samples with each label is decided via random selection, but here we cannot apply this selection strategy. We need to make the numbers of data samples belonging to each label have differences, otherwise we cannot evaluate the performance of Whole Determination. To start with, we should figure out the connection between the magnitude of ratio difference and the proportion difference of labels. We conduct experiments by changing the number of samples belonging to a certain label and record the corresponding ratio difference. The results are presented in Figure~\ref{fig_wd_ratio}. As shown in the figure, obvious ratio difference can be observed when there is a four-fold difference in the number of samples owned by the two labels. As a result, we divide the whole labels into 3 groups randomly, and ensure that each label in the first group can be allocated with $Q$ data samples, the second group can only acquire $Q/4$, and the last group just get $Q/5$ samples. These groups will be used to train learning model, and what we want is to evaluate if our approach can detect this composition proportion. 

\subsubsection{Results}
We conduct the experiments on all datasets. For each dataset, we train it for 20 times and launch Whole Determination attack in every training. The middle stage of training is defined as the round when the testing accuracy is approximately $50\%$, while the late stage is when the testing accuracy exceeds $85\%$. Clearly, the dataset allocation is different in each training process because of the random allocation. We also use success rate as our metric, and only consider the attack successful when the clustering results are exactly the same as the data allocation before training, including the number of clusters and the specific labels in each cluster; otherwise we regard it as a failure.

The results are presented in Table~\ref{table_5}. We can see that the average success rate is very high (almost 95\%), which shows that Whole Determination attack can be effective under such circumstances. The success rate of middle stage is a little worse than that of late stage. We think the reason could be that the exploration direction of gradient is relatively more random in the middle stage of training.

\begin{table}[!t]
\centering
\caption{The results of Whole Determination on experiment datasets}
\label{table_5}
\begin{tabular}{|c|c|c|c|}
\hline
Dataset                  & Model    & \multicolumn{2}{c|}{Success Rate(\%)} \\ \hline
\multicolumn{2}{|c|}{Stages}        & Middle             & Late             \\ \hline
\multirow{2}{*}{MNIST}   & MLP      & 95                 & 100              \\ \cline{2-4} 
                         & CNN      & 95                 & 100              \\ \hline
\multirow{2}{*}{CIFAR10} & LeNet5      & 90                 & 95               \\ \cline{2-4} 
                         & Resnet18 & 95                 & 100              \\ \hline
Fer2013                  & Resnet18 & 95                 & 95               \\ \hline
HAM10000                 & Resnet18 & 95                 & 95               \\ \hline
\end{tabular}
\end{table}

\section{Discussion}
\subsection{Network Layers}
It could be asked why we consider the output layer and whether similar phenomenon exists in other layers, e.g., hidden layers. 
The main task of the front network layers is typically to extract and filter the features of training data, which means the objects to be processed are various features. The emergence of a particular feature leads to corresponding gradient updates, while its absence has no influence~\cite{ateniese2013hacking}. It is interesting to note that a certain label usually possesses many different features, in other words, it is an unity of multiple features. And different labels possibly have the same features, e.g., cats and dogs have similar fur features. What is more, in some special neural networks, the features embedded in front layers are not explainable or interpretable to human analysts practically, especially the cases in convolution operation. These characteristics make the front layers not applicable to our case, where we want to obtain the quantity information about training labels. However, in case we cannot access the output layer but only several front layers, we could try to apply some Explainable Machine Learning techniques to extract the key features of each class, like LIME~\cite{ribeiro2016should} for linear classifiers and LEMNA~\cite{guo2018lemna} for deep neural networks. Explainable Machine Learning aims at figuring out why a particular input sample is clustered into its corresponding label and obtaining some relatively interpretable reasons for users, especially for debuggers and computer security practitioners. And these reasons, namely the key features, can be served as the identifications for different labels, and our method is possibly able to build on them.

\subsection{Defense}
The three attacks proposed by us share similarity to property inference attack (albeit our focus switches from the existential proof to quantity information). Thus, some defenses designed for property inference may be leveraged to mitigate our attacks. Here we discuss two possible defenses.

\subsubsection{Compress the gradient updates}
As mentioned by \cite{shokri2015privacy}, there is no need to share the whole network parameters in collaborative learning. In other words, compressing or distilling the significant neuron updates can also make the global model converge and achieve great performance. As for our attacks, the adversary does not need to observe any individual updates, and what he needs is only the global model parameters, hence there is some possible impact to the attacks if the participants cannot acquire the whole global model (it can only acquire part of it). 

We try to simulate such defense in our experiment setting, and we choose to keep the weights whose gradient updates are relatively greater and make other weights invisible to participants. Specifically, if we set the compression rate (CR) in advance, each client will select top CR of inputting weight updates to be uploaded to the server. This simple compression operation might make the convergence of global model slower to some extent, but the influence to performance of the model is not great. And under such circumstances, we conduct some experiments on Quantity Inference under the default setting (MNIST on CNN). The results can be seen in Table~\ref{defense}. 

\begin{table}[!t]
\centering
\caption{The results of compression defenses on MNIST}
\label{defense}
\begin{tabular}{|c|c|c|}
\hline
Compression Rate & Success Rate(\%) & Aborting Rate(\%) \\ \hline
0                & 96               & 0                 \\ \hline
0.1              & 95               & 3                 \\ \hline
0.2              & 96               & 10                \\ \hline
0.3              & 97               & 20                \\ \hline
0.4              & 94               & 28                \\ \hline
0.5              & 95               & 37                \\ \hline
\end{tabular}
\end{table}

We launch Quantity Inference for 100 rounds under each CR setting. From the results, we can see that the success rate of our attack is not impacted by such defense. However, the aborting rate increases a lot since Quantity Inference is designed to abort the round whose corresponding standard deviation of $x$'s (Sec.~\ref{qi}) is high. Thus, the aborting operation indirectly makes this attack less effective.

\subsubsection{Dropout}
Another possible defense is adding Dropout Layer to the neural networks, which is also an effective approach to mitigate the overfitting phenomenon. Since its random removal of the features in training process, the dropout operation may make the gradient updates of clients more different from each other, which possibly poses some impact to our attacks. However in our initial evaluation experiments, especially for MNIST dataset, both MLP and CNN (Sec.~\ref{dataset}) have dropout layers, and the success rate of our three attacks is still extremely high. Thus, we are not sure whether the dropout technique is able to defense our attack methods and plan to conduct a deeper analysis in the future.

\section{Related Work}\label{sec:related}

\subsection{Privacy-preserving Federated Learning}
Federated learning is an evolution form of distributed learning, and it enables training data stay locally while a collaborative global model can be learned. Existing work with the considering of privacy can be classified into Differential Privacy mechanism (DP) and secure Multi-Party Computation (MPC). \citet{geyer2017differentially} stand on the perspective of the client and realize differential privacy protection by adding Gaussian noise to the local updates under the setting where there are a large number of participants, and similar work can also be seen in~\cite{mcmahan2017learning}. \citet{hamm2016learning} apply knowledge transfer techniques to aggregate multiple models trained on individual devices with DP guarantee.

\citet{bonawitz2017practical} design secure multi-party aggregation techniques, pertinent for federated learning, to enable participants to encrypt themselves so that the central server cannot observe individual gradient updates in the plain form and only do the aggregation operation. \citet{mohassel2017secureml} enable two servers to train a global model with multi-party encrypted data, and the training process is protected by MPC techniques.

\subsection{Inference Attack}
Different types of inference attacks in collaborative setting emerge frequently. \citet{hitaj2017deep} create a GAN structure to imitate the output probability distributions and use reverse learning to infer the training data. \citet{hayes2019logan} note the privacy leakage in the scenario of machine-learning-as-a-service application and also train several GANs to detect overfitting characteristics of input-output pairs. \citet{truex2019demystifying} propose a membership inference threat on the surface of FL, but they assume FL is under machine-learning-as-a-service application and adversaries hold the ability that sniffing output probability distributions of all other clients rather than model parameter updates, which we think it is not an inference attack towards the standard FL. \citet{melis2018exploiting} lay emphasis on the unintended feature leakage in collaborative learning setting by training a shadow attack model to infer information about training data, and the threat model have been simplified to some extent. Different from above works, \citet{wang2019beyond} assume that the aggregation server in FL is malicious, and they combine the main work of global model, identity distinguishing and traditional authentication task to form the mixed discriminator of a GAN that is able to track particular victims and reconstruct their private training data. Much similar to the aggregation feature of FL, in aggregated location field, \citet{pyrgelis2017knock} use a challenge game to distinguish the victims with other participates and then track the location information of particular victims. 

There are also some relative work about property inference attack for traditional machine learning, in both white-box and black-box settings. For instance, under black-box circumstance, \citet{salem2019updates} use GAN to achieve reconstruction and then sniff the information about training data between different versions of learning model based on updates of output results. \citet{ateniese2013hacking} use different property of training data to obtain several meta-models, and combine these meta-models to sense the existence of a particular label. \citet{ganju2018property} construct an inference attack towards fully-connected neural networks, and they realize it by applying post-training techniques to a white-box model.

\subsection{Other Attacks and Defenses}
\noindent\textbf{Attack.}
Federated learning is a fertile research field for security problems, and there have been several other interesting attacks recently. \citet{bagdasaryan2018backdoor} create a type of backdoor approach in FL setting, which can pose the backdoor threat after only a few rounds of attack with high target class accuracy. \citet{baruch2019little} propose a poisoning attack whose impact is profound and it can escape prevalent abnormal detection. They realize it by splitting abnormal parts in a few neurons to a large number of neurons, and they also investigate the capacity scope of current abnormal detection approaches on the degree of abnormality. \citet{bhagoji2018analyzing} explore the threat of model poisoning attacks on FL launched by a single, non-colluding malicious agent, where the adversarial objective is to cause the model to mis-classify a set of chosen inputs with high confidence.

\medskip\noindent\textbf{Defense.}
\citet{shen2016uror} apply the clustering operation to individual parameter updates before aggregation to detect malicious participants in distributed learning setting. \citet{blanchard2017machine} use Euclidean Distance to measure the contribution of clients to global model, and design a selection strategy to tolerate the gradient contribution from Byzantine attackers. \citet{fung2018mitigating} present the impact of sybils attack in FL and design a detection algorithm by comparing the cosine similarity between gradient updates.

\section{Conclusion}
In this paper, we proposed three original inference attacks against federated learning. The attack target includes the quantity composition proportion of training labels, a new consideration in FL security. Specifically, \textbf{Class Sniffing} can detect the existence of a particular label in a single training round; \textbf{Quantity Inference} is able to determine how many clients own a certain label from the perspective of a single iteration; and finally, \textbf{Whole Determination} aims to infer the quantity information among different labels for the whole training process. All of them work in a passive way, and they will not impose any influence to the whole FL structure, hence it is difficult for the prevalent intrusion detection techniques to detect our attacks. Besides, all three attacks do not require the observation of any individual gradient updates from participants, which enables the attackers to apply them in more practical scenarios.

We have conducted extensive experiments that demonstrate the effectiveness of our attacks, with evaluation settings as practical as we can. All three attacks are shown to be very effective, with their success rates staying at a relative high level (typically around $95\%$). Moreover, we also investigated the impact of major hyper-parameters, e.g., batch size, local epochs and the overall number of participants. The results demonstrate broad applicability of our approaches.



%
%
%
\bibliographystyle{unsrtnat}
\bibliography{references}  

\begin{thebibliography}{49}
\providecommand{\natexlab}[1]{#1}
\providecommand{\url}[1]{\texttt{#1}}
\expandafter\ifx\csname urlstyle\endcsname\relax
  \providecommand{\doi}[1]{doi: #1}\else
  \providecommand{\doi}{doi: \begingroup \urlstyle{rm}\Url}\fi

\bibitem[Anguita et~al.(2013)Anguita, Ghio, Oneto, Parra, and
  Reyes-Ortiz]{anguita2013public}
Davide Anguita, Alessandro Ghio, Luca Oneto, Xavier Parra, and Jorge~Luis
  Reyes-Ortiz.
\newblock A public domain dataset for human activity recognition using
  smartphones.
\newblock In \emph{Esann}, 2013.

\bibitem[Hard et~al.(2018)Hard, Rao, Mathews, Beaufays, Augenstein, Eichner,
  Kiddon, and Ramage]{hard2018federated}
Andrew Hard, Kanishka Rao, Rajiv Mathews, Fran{\c{c}}oise Beaufays, Sean
  Augenstein, Hubert Eichner, Chlo{\'e} Kiddon, and Daniel Ramage.
\newblock Federated learning for mobile keyboard prediction.
\newblock \emph{arXiv preprint arXiv:1811.03604}, 2018.

\bibitem[Ramaswamy et~al.(2019)Ramaswamy, Mathews, Rao, and
  Beaufays]{ramaswamy2019federated}
Swaroop Ramaswamy, Rajiv Mathews, Kanishka Rao, and Fran{\c{c}}oise Beaufays.
\newblock Federated learning for emoji prediction in a mobile keyboard.
\newblock \emph{arXiv preprint arXiv:1906.04329}, 2019.

\bibitem[Pantelopoulos and Bourbakis(2009)]{pantelopoulos2009survey}
Alexandros Pantelopoulos and Nikolaos~G Bourbakis.
\newblock A survey on wearable sensor-based systems for health monitoring and
  prognosis.
\newblock \emph{IEEE Transactions on Systems, Man, and Cybernetics, Part C
  (Applications and Reviews)}, 40\penalty0 (1):\penalty0 1--12, 2009.

\bibitem[Nguyen et~al.(2018)Nguyen, Marchal, Miettinen, Fereidooni, Asokan, and
  Sadeghi]{nguyen2018d}
Thien~Duc Nguyen, Samuel Marchal, Markus Miettinen, Hossein Fereidooni,
  N~Asokan, and Ahmad-Reza Sadeghi.
\newblock D$\backslash$" iot: A federated self-learning anomaly detection
  system for iot.
\newblock \emph{arXiv preprint arXiv:1804.07474}, 2018.

\bibitem[Samarakoon et~al.(2018{\natexlab{a}})Samarakoon, Bennis, Saady, and
  Debbah]{samarakoon2018distributed}
Sumudu Samarakoon, Mehdi Bennis, Walid Saady, and Merouane Debbah.
\newblock Distributed federated learning for ultra-reliable low-latency
  vehicular communications.
\newblock \emph{arXiv preprint arXiv:1807.08127}, 2018{\natexlab{a}}.

\bibitem[Samarakoon et~al.(2018{\natexlab{b}})Samarakoon, Bennis, Saad, and
  Debbah]{samarakoon2018federated}
Sumudu Samarakoon, Mehdi Bennis, Walid Saad, and Merouane Debbah.
\newblock Federated learning for ultra-reliable low-latency v2v communications.
\newblock In \emph{2018 IEEE Global Communications Conference (GLOBECOM)},
  pages 1--7. IEEE, 2018{\natexlab{b}}.

\bibitem[Chilimbi et~al.(2014)Chilimbi, Suzue, Apacible, and
  Kalyanaraman]{chilimbi2014project}
Trishul Chilimbi, Yutaka Suzue, Johnson Apacible, and Karthik Kalyanaraman.
\newblock Project adam: Building an efficient and scalable deep learning
  training system.
\newblock In \emph{11th $\{$USENIX$\}$ Symposium on Operating Systems Design
  and Implementation ($\{$OSDI$\}$ 14)}, pages 571--582, 2014.

\bibitem[Dean et~al.(2012)Dean, Corrado, Monga, Chen, Devin, Mao, Senior,
  Tucker, Yang, Le, et~al.]{dean2012large}
Jeffrey Dean, Greg Corrado, Rajat Monga, Kai Chen, Matthieu Devin, Mark Mao,
  Andrew Senior, Paul Tucker, Ke~Yang, Quoc~V Le, et~al.
\newblock Large scale distributed deep networks.
\newblock In \emph{Advances in neural information processing systems}, pages
  1223--1231, 2012.

\bibitem[Lin et~al.(2017{\natexlab{a}})Lin, Han, Mao, Wang, and
  Dally]{lin2017deep}
Yujun Lin, Song Han, Huizi Mao, Yu~Wang, and William~J Dally.
\newblock Deep gradient compression: Reducing the communication bandwidth for
  distributed training.
\newblock \emph{arXiv preprint arXiv:1712.01887}, 2017{\natexlab{a}}.

\bibitem[Moritz et~al.(2015)Moritz, Nishihara, Stoica, and
  Jordan]{moritz2015sparknet}
Philipp Moritz, Robert Nishihara, Ion Stoica, and Michael~I Jordan.
\newblock Sparknet: Training deep networks in spark.
\newblock \emph{arXiv preprint arXiv:1511.06051}, 2015.

\bibitem[Xing et~al.(2015)Xing, Ho, Dai, Kim, Wei, Lee, Zheng, Xie, Kumar, and
  Yu]{xing2015petuum}
Eric~P Xing, Qirong Ho, Wei Dai, Jin~Kyu Kim, Jinliang Wei, Seunghak Lee, Xun
  Zheng, Pengtao Xie, Abhimanu Kumar, and Yaoliang Yu.
\newblock Petuum: A new platform for distributed machine learning on big data.
\newblock \emph{IEEE Transactions on Big Data}, 1\penalty0 (2):\penalty0
  49--67, 2015.

\bibitem[Zinkevich et~al.(2010)Zinkevich, Weimer, Li, and
  Smola]{zinkevich2010parallelized}
Martin Zinkevich, Markus Weimer, Lihong Li, and Alex~J Smola.
\newblock Parallelized stochastic gradient descent.
\newblock In \emph{Advances in neural information processing systems}, pages
  2595--2603, 2010.

\bibitem[Bagdasaryan et~al.(2018)Bagdasaryan, Veit, Hua, Estrin, and
  Shmatikov]{bagdasaryan2018backdoor}
Eugene Bagdasaryan, Andreas Veit, Yiqing Hua, Deborah Estrin, and Vitaly
  Shmatikov.
\newblock How to backdoor federated learning.
\newblock \emph{arXiv preprint arXiv:1807.00459}, 2018.

\bibitem[Baruch et~al.(2019)Baruch, Baruch, and Goldberg]{baruch2019little}
Moran Baruch, Gilad Baruch, and Yoav Goldberg.
\newblock A little is enough: Circumventing defenses for distributed learning.
\newblock \emph{arXiv preprint arXiv:1902.06156}, 2019.

\bibitem[Fung et~al.(2018)Fung, Yoon, and Beschastnikh]{fung2018mitigating}
Clement Fung, Chris~JM Yoon, and Ivan Beschastnikh.
\newblock Mitigating sybils in federated learning poisoning.
\newblock \emph{arXiv preprint arXiv:1808.04866}, 2018.

\bibitem[Bhagoji et~al.(2018)Bhagoji, Chakraborty, Mittal, and
  Calo]{bhagoji2018analyzing}
Arjun~Nitin Bhagoji, Supriyo Chakraborty, Prateek Mittal, and Seraphin Calo.
\newblock Analyzing federated learning through an adversarial lens.
\newblock \emph{arXiv preprint arXiv:1811.12470}, 2018.

\bibitem[Mahloujifar et~al.(2018)Mahloujifar, Mahmoody, and
  Mohammed]{mahloujifar2018multi}
Saeed Mahloujifar, Mohammad Mahmoody, and Ameer Mohammed.
\newblock Multi-party poisoning through generalized $ p $-tampering.
\newblock \emph{arXiv preprint arXiv:1809.03474}, 2018.

\bibitem[Melis et~al.(2018)Melis, Song, De~Cristofaro, and
  Shmatikov]{melis2018exploiting}
Luca Melis, Congzheng Song, Emiliano De~Cristofaro, and Vitaly Shmatikov.
\newblock Exploiting unintended feature leakage in collaborative learning.
\newblock \emph{arXiv preprint arXiv:1805.04049}, 2018.

\bibitem[Truex et~al.(2019)Truex, Liu, Gursoy, Yu, and
  Wei]{truex2019demystifying}
Stacey Truex, Ling Liu, Mehmet~Emre Gursoy, Lei Yu, and Wenqi Wei.
\newblock Demystifying membership inference attacks in machine learning as a
  service.
\newblock \emph{IEEE Transactions on Services Computing}, 2019.

\bibitem[Pyrgelis et~al.(2017)Pyrgelis, Troncoso, and
  De~Cristofaro]{pyrgelis2017knock}
Apostolos Pyrgelis, Carmela Troncoso, and Emiliano De~Cristofaro.
\newblock Knock knock, who's there? membership inference on aggregate location
  data.
\newblock \emph{arXiv preprint arXiv:1708.06145}, 2017.

\bibitem[Nasr et~al.(2018)Nasr, Shokri, and Houmansadr]{nasr2018comprehensive}
Milad Nasr, Reza Shokri, and Amir Houmansadr.
\newblock Comprehensive privacy analysis of deep learning: Stand-alone and
  federated learning under passive and active white-box inference attacks.
\newblock \emph{arXiv preprint arXiv:1812.00910}, 2018.

\bibitem[Hitaj et~al.(2017)Hitaj, Ateniese, and Perez-Cruz]{hitaj2017deep}
Briland Hitaj, Giuseppe Ateniese, and Fernando Perez-Cruz.
\newblock Deep models under the gan: information leakage from collaborative
  deep learning.
\newblock In \emph{Proceedings of the 2017 ACM SIGSAC Conference on Computer
  and Communications Security}, pages 603--618. ACM, 2017.

\bibitem[Salem et~al.(2019)Salem, Bhattacharya, Backes, Fritz, and
  Zhang]{salem2019updates}
Ahmed Salem, Apratim Bhattacharya, Michael Backes, Mario Fritz, and Yang Zhang.
\newblock Updates-leak: Data set inference and reconstruction attacks in online
  learning.
\newblock \emph{arXiv preprint arXiv:1904.01067}, 2019.

\bibitem[Hayes et~al.(2019)Hayes, Melis, Danezis, and
  De~Cristofaro]{hayes2019logan}
Jamie Hayes, Luca Melis, George Danezis, and Emiliano De~Cristofaro.
\newblock Logan: Membership inference attacks against generative models.
\newblock \emph{Proceedings on Privacy Enhancing Technologies}, 2019\penalty0
  (1):\penalty0 133--152, 2019.

\bibitem[Wang et~al.(2019)Wang, Song, Zhang, Song, Wang, and
  Qi]{wang2019beyond}
Zhibo Wang, Mengkai Song, Zhifei Zhang, Yang Song, Qian Wang, and Hairong Qi.
\newblock Beyond inferring class representatives: User-level privacy leakage
  from federated learning.
\newblock In \emph{IEEE INFOCOM 2019-IEEE Conference on Computer
  Communications}, pages 2512--2520. IEEE, 2019.

\bibitem[Aono et~al.(2017)Aono, Hayashi, Wang, Moriai, et~al.]{aono2017privacy}
Yoshinori Aono, Takuya Hayashi, Lihua Wang, Shiho Moriai, et~al.
\newblock Privacy-preserving deep learning: Revisited and enhanced.
\newblock In \emph{International Conference on Applications and Techniques in
  Information Security}, pages 100--110. Springer, 2017.

\bibitem[Bonawitz et~al.(2017)Bonawitz, Ivanov, Kreuter, Marcedone, McMahan,
  Patel, Ramage, Segal, and Seth]{bonawitz2017practical}
Keith Bonawitz, Vladimir Ivanov, Ben Kreuter, Antonio Marcedone, H~Brendan
  McMahan, Sarvar Patel, Daniel Ramage, Aaron Segal, and Karn Seth.
\newblock Practical secure aggregation for privacy-preserving machine learning.
\newblock In \emph{Proceedings of the 2017 ACM SIGSAC Conference on Computer
  and Communications Security}, pages 1175--1191. ACM, 2017.

\bibitem[Geyer et~al.(2017)Geyer, Klein, and Nabi]{geyer2017differentially}
Robin~C Geyer, Tassilo Klein, and Moin Nabi.
\newblock Differentially private federated learning: A client level
  perspective.
\newblock \emph{arXiv preprint arXiv:1712.07557}, 2017.

\bibitem[McMahan et~al.(2017)McMahan, Ramage, Talwar, and
  Zhang]{mcmahan2017learning}
H~Brendan McMahan, Daniel Ramage, Kunal Talwar, and Li~Zhang.
\newblock Learning differentially private language models without losing
  accuracy.
\newblock \emph{arXiv preprint arXiv:1710.06963}, 2017.

\bibitem[Huang et~al.(2018)Huang, Yin, Fu, Zhang, Deng, and
  Liu]{huang2018loadaboost}
Li~Huang, Yifeng Yin, Zeng Fu, Shifa Zhang, Hao Deng, and Dianbo Liu.
\newblock Loadaboost: Loss-based adaboost federated machine learning on medical
  data.
\newblock \emph{arXiv preprint arXiv:1811.12629}, 2018.

\bibitem[Ateniese et~al.(2013)Ateniese, Felici, Mancini, Spognardi, Villani,
  and Vitali]{ateniese2013hacking}
Giuseppe Ateniese, Giovanni Felici, Luigi~V Mancini, Angelo Spognardi, Antonio
  Villani, and Domenico Vitali.
\newblock Hacking smart machines with smarter ones: How to extract meaningful
  data from machine learning classifiers.
\newblock \emph{arXiv preprint arXiv:1306.4447}, 2013.

\bibitem[Lin et~al.(2017{\natexlab{b}})Lin, Goyal, Girshick, He, and
  Doll{\'a}r]{lin2017focal}
Tsung-Yi Lin, Priya Goyal, Ross Girshick, Kaiming He, and Piotr Doll{\'a}r.
\newblock Focal loss for dense object detection.
\newblock In \emph{Proceedings of the IEEE international conference on computer
  vision}, pages 2980--2988, 2017{\natexlab{b}}.

\bibitem[Paszke et~al.(2017)Paszke, Gross, Chintala, and
  Chanan]{paszke2017pytorch}
Adam Paszke, Sam Gross, Soumith Chintala, and Gregory Chanan.
\newblock Pytorch.
\newblock \emph{Computer software. Vers. 0.3}, 1, 2017.

\bibitem[Pedregosa et~al.(2011)Pedregosa, Varoquaux, Gramfort, Michel, Thirion,
  Grisel, Blondel, Prettenhofer, Weiss, Dubourg, et~al.]{pedregosa2011scikit}
Fabian Pedregosa, Ga{\"e}l Varoquaux, Alexandre Gramfort, Vincent Michel,
  Bertrand Thirion, Olivier Grisel, Mathieu Blondel, Peter Prettenhofer, Ron
  Weiss, Vincent Dubourg, et~al.
\newblock Scikit-learn: Machine learning in python.
\newblock \emph{Journal of machine learning research}, 12\penalty0
  (Oct):\penalty0 2825--2830, 2011.

\bibitem[Kone{\v{c}}n{\`y} et~al.(2016)Kone{\v{c}}n{\`y}, McMahan, Yu,
  Richt{\'a}rik, Suresh, and Bacon]{konevcny2016federated}
Jakub Kone{\v{c}}n{\`y}, H~Brendan McMahan, Felix~X Yu, Peter Richt{\'a}rik,
  Ananda~Theertha Suresh, and Dave Bacon.
\newblock Federated learning: Strategies for improving communication
  efficiency.
\newblock \emph{arXiv preprint arXiv:1610.05492}, 2016.

\bibitem[McMahan et~al.(2016)McMahan, Moore, Ramage, Hampson,
  et~al.]{mcmahan2016communication}
H~Brendan McMahan, Eider Moore, Daniel Ramage, Seth Hampson, et~al.
\newblock Communication-efficient learning of deep networks from decentralized
  data.
\newblock \emph{arXiv preprint arXiv:1602.05629}, 2016.

\bibitem[LeCun et~al.(1998)LeCun, Bottou, Bengio, Haffner,
  et~al.]{lecun1998gradient}
Yann LeCun, L{\'e}on Bottou, Yoshua Bengio, Patrick Haffner, et~al.
\newblock Gradient-based learning applied to document recognition.
\newblock \emph{Proceedings of the IEEE}, 86\penalty0 (11):\penalty0
  2278--2324, 1998.

\bibitem[He et~al.(2016)He, Zhang, Ren, and Sun]{he2016deep}
Kaiming He, Xiangyu Zhang, Shaoqing Ren, and Jian Sun.
\newblock Deep residual learning for image recognition.
\newblock In \emph{Proceedings of the IEEE conference on computer vision and
  pattern recognition}, pages 770--778, 2016.

\bibitem[Goodfellow et~al.(2013)Goodfellow, Erhan, Carrier, Courville, Mirza,
  Hamner, Cukierski, Tang, Thaler, Lee, et~al.]{goodfellow2013challenges}
Ian~J Goodfellow, Dumitru Erhan, Pierre~Luc Carrier, Aaron Courville, Mehdi
  Mirza, Ben Hamner, Will Cukierski, Yichuan Tang, David Thaler, Dong-Hyun Lee,
  et~al.
\newblock Challenges in representation learning: A report on three machine
  learning contests.
\newblock In \emph{International Conference on Neural Information Processing},
  pages 117--124. Springer, 2013.

\bibitem[Tschandl et~al.(2018)Tschandl, Rosendahl, and
  Kittler]{tschandl2018ham10000}
Philipp Tschandl, Cliff Rosendahl, and Harald Kittler.
\newblock The ham10000 dataset, a large collection of multi-source
  dermatoscopic images of common pigmented skin lesions.
\newblock \emph{Scientific data}, 5:\penalty0 180161, 2018.

\bibitem[Ribeiro et~al.(2016)Ribeiro, Singh, and Guestrin]{ribeiro2016should}
Marco~Tulio Ribeiro, Sameer Singh, and Carlos Guestrin.
\newblock Why should i trust you?: Explaining the predictions of any
  classifier.
\newblock In \emph{Proceedings of the 22nd ACM SIGKDD international conference
  on knowledge discovery and data mining}, pages 1135--1144. ACM, 2016.

\bibitem[Guo et~al.(2018)Guo, Mu, Xu, Su, Wang, and Xing]{guo2018lemna}
Wenbo Guo, Dongliang Mu, Jun Xu, Purui Su, Gang Wang, and Xinyu Xing.
\newblock Lemna: Explaining deep learning based security applications.
\newblock In \emph{Proceedings of the 2018 ACM SIGSAC Conference on Computer
  and Communications Security}, pages 364--379. ACM, 2018.

\bibitem[Shokri and Shmatikov(2015)]{shokri2015privacy}
Reza Shokri and Vitaly Shmatikov.
\newblock Privacy-preserving deep learning.
\newblock In \emph{Proceedings of the 22nd ACM SIGSAC conference on computer
  and communications security}, pages 1310--1321. ACM, 2015.

\bibitem[Hamm et~al.(2016)Hamm, Cao, and Belkin]{hamm2016learning}
Jihun Hamm, Yingjun Cao, and Mikhail Belkin.
\newblock Learning privately from multiparty data.
\newblock In \emph{International Conference on Machine Learning}, pages
  555--563, 2016.

\bibitem[Mohassel and Zhang(2017)]{mohassel2017secureml}
Payman Mohassel and Yupeng Zhang.
\newblock Secureml: A system for scalable privacy-preserving machine learning.
\newblock In \emph{2017 IEEE Symposium on Security and Privacy (SP)}, pages
  19--38. IEEE, 2017.

\bibitem[Ganju et~al.(2018)Ganju, Wang, Yang, Gunter, and
  Borisov]{ganju2018property}
Karan Ganju, Qi~Wang, Wei Yang, Carl~A Gunter, and Nikita Borisov.
\newblock Property inference attacks on fully connected neural networks using
  permutation invariant representations.
\newblock In \emph{Proceedings of the 2018 ACM SIGSAC Conference on Computer
  and Communications Security}, pages 619--633. ACM, 2018.

\bibitem[Shen et~al.(2016)Shen, Tople, and Saxena]{shen2016uror}
Shiqi Shen, Shruti Tople, and Prateek Saxena.
\newblock A uror: defending against poisoning attacks in collaborative deep
  learning systems.
\newblock In \emph{Proceedings of the 32nd Annual Conference on Computer
  Security Applications}, pages 508--519. ACM, 2016.

\bibitem[Blanchard et~al.(2017)Blanchard, Guerraoui, Stainer,
  et~al.]{blanchard2017machine}
Peva Blanchard, Rachid Guerraoui, Julien Stainer, et~al.
\newblock Machine learning with adversaries: Byzantine tolerant gradient
  descent.
\newblock In \emph{Advances in Neural Information Processing Systems}, pages
  119--129, 2017.

\end{thebibliography}

\end{document}